\documentclass{article}

\usepackage[preprint]{neurips_2026}

\usepackage[utf8]{inputenc} 
\usepackage[T1]{fontenc}    
\usepackage{hyperref}       
\usepackage{url}            
\usepackage{booktabs}       
\usepackage{amsfonts}       
\usepackage{nicefrac}       
\usepackage{microtype}      
\usepackage{xcolor}         
\usepackage{amssymb}
\usepackage{graphicx}
\usepackage{amsmath}
\usepackage{multirow}
\usepackage{wrapfig}
\usepackage{multicol}
\usepackage{extarrows}
\usepackage{lipsum}
\usepackage{subcaption}
\usepackage{siunitx}

\newcolumntype{R}[1]{>{\raggedleft\arraybackslash}p{#1}}

\title{Small Models, Strong Priors: Architectural Inductive Bias for Parameter-Efficient Neural PDE Solvers}

\author{%
  Shyam Sankaran \\
  Department of Mechanical Engineering and Applied Mechanics, University of Pennsylvania \\
  \texttt{shyamss@seas.upenn.edu} \\
  \And
  Hanwen Wang \\
  Graduate Program in Applied Mathematics and Computational Science, University of Pennsylvania \\
  \texttt{wangh19@sas.upenn.edu} \\
  \And
  Paris Perdikaris \\
  Department of Mechanical Engineering and Applied Mechanics, University of Pennsylvania \\
  \texttt{pgp@seas.upenn.edu}  \\
}

\begin{document}

\maketitle

\begin{abstract}
Neural PDE solvers have followed the scaling trajectory of vision and language, with recent foundation models reaching billions of parameters. We argue that scale is a poor substitute for architectural inductive bias in this domain: structured priors deliver outsized parameter efficiency, and the pattern of where they succeed and fail is itself informative about what they capture. We instantiate this argument in \emph{WaveLiT}, an architecture combining a discrete wavelet transform for lossless multi-resolution tokenization, an augmented linear attention block, a shared-weight multiscale feature pyramid, and a wavelet-domain auxiliary loss. Bespoke 1--10M-parameter WaveLiT models compete with foundation models of 100--1000$\times$ their size across eight TheWell benchmarks, with the largest gains on wave- and acoustic-dominated benchmarks where the wavelet-multiscale prior fits the dominant dynamical structure and small per-step errors do not compound geometrically under rollout.
Trained jointly across all eight benchmarks, a 10M-parameter foundation variant exhibits a structured, physically interpretable transfer pattern -- strongest where the wavelet-multiscale prior matches the dynamics, weakest on chaotic advection-dominated flows. The entire pipeline trains on a single GPU. The results suggest that small-model PDE performance is shaped by architectural inductive bias rather than scale, and that the structure of a prior's failures is a useful empirical signal about its content.
\end{abstract}

\section{Introduction}
\label{introduction}

\begin{wrapfigure}[9]{r}{0.45\columnwidth} 
    \centering
\vspace{-1.7cm}
\includegraphics[width=\linewidth]{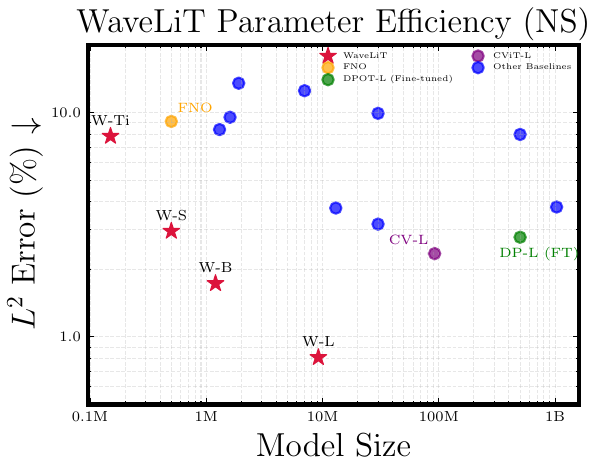}
\vspace{-0.8cm}
\caption{\small{Parameter efficiency on PDEArena Navier-Stokes \cite{gupta2022towards}. Each point is a model; lower-left is better. WaveLiT (red stars) outperforms models with $100\times$ more parameters.}}
\label{fig:param_efficiency_intro_wrap}
\end{wrapfigure}

Neural surrogates for partial differential equations are increasingly central to scientific computing, with applications spanning weather forecasting \cite{bodnar2024aurora}, fluid dynamics, and materials science \cite{wang2024micrometer}. The promise is straightforward: where classical solvers cost hours per simulation, learned surrogates cost milliseconds. Realizing this promise at the accuracy required for scientific use has, in current practice, driven the field toward ever-larger models.

Following the scaling regimes established in language\cite{kaplan2020scaling} and vision \cite{zhai2022scaling}, neural PDE solvers have moved from thousands of parameters to billions. Recent foundation models such as PhysiX (4.5B) \cite{nguyen2025physix}, DPOT (500M) \cite{hao2024dpot}, Poseidon (629M) \cite{herde2024poseidon}, and Walrus (1.2B) \cite{mccabe2025walrus} pursue general-purpose simulators by combining massive parameter counts with multi-task training across diverse PDE families. Training these models requires hundreds of GPU-hours and infrastructure that is inaccessible to most research groups.

At the same time, the broader ML community is increasingly recognizing that scale is not the only axis of progress. Compact language models have emerged as efficient backbones for downstream systems \cite{abdin2024phi, belcak2025small}, demonstrating that careful architectural and data choices can substitute for raw parameter count. 
A complementary direction in reasoning has shown that small networks with appropriate inductive structure -- HRM at 27M parameters \cite{wang2025hierarchical} and TRM at 7M \cite{jolicoeur2025less} -- can be competitive on tasks with exploitable combinatorial structure.
Neural PDE solvers occupy an analogous setting: the underlying physics imposes locality, regularity, and multi-scale structure that should, in principle, be exploitable by smaller models, yet architectural defaults in this field have not been re-examined under this lens.

Most current PDE models inherit their design from vision transformers, adopting defaults that were never optimized for physical systems. Coarse 8$\times$8 or 16$\times$16 patch tokenization discards sub-patch physics before the first attention layer, and quadratic self-attention forces this coarse tokenization by making fine-grained alternatives prohibitively expensive. These defaults interact in a vicious cycle: quadratic attention demands fewer tokens, which demands coarser patches, which destroys fine-scale physics, which demands more parameters to compensate.

We introduce \emph{WaveLiT}, an integrated architecture that addresses each of these defaults. A discrete wavelet transform replaces learned patch embedding with parameter-free, lossless multi-resolution tokenization. A linear attention block with $\mathcal{O}(N)$ complexity replaces quadratic self-attention, building on Mamba-Inspired Linear Attention~\cite{han2024demystify} and the test-time-regression view of attention~\cite{von2025mesanet, behrouz2025s} adapted to the bidirectional spatial setting. A shared-weight multiscale feature pyramid processes tokens at multiple resolutions without adding parameters, and a wavelet-domain auxiliary loss encourages spectral fidelity across scales.

These choices compound into a parameter-efficient frontier (Figure~\ref{fig:param_efficiency_intro_wrap}). On PDEArena Navier-Stokes \cite{gupta2022towards}, WaveLiT matches or exceeds FNO \cite{li2020fourier}, CViT-L \cite{wang2024cvit}, and DPOT-L \cite{hao2024dpot} at $10$--$100\times$ fewer parameters; 
on TheWell \cite{ohana2024well}, bespoke 1--10M-parameter models compete with foundation-model baselines (MPP-AViT-L, Poseidon-L, DPOT-H, Walrus) of $100$--$1000\times$ their size, with the largest gains on wave- and acoustic-dominated benchmarks. To probe whether these architectural priors survive cross-family training, we use the foundation-model regime as a diagnostic instrument: a single 10M-parameter model trained jointly on all eight TheWell benchmarks exhibits a structured, physically interpretable transfer pattern -- strongest where the wavelet-multiscale prior matches the dynamics, weakest on chaotic advection-dominated flows. The entire pipeline trains on a single GPU. We do not claim global SOTA against billion-parameter specialists; we claim that the recipe defines an efficient frontier, and that the structure of its successes and failures provides evidence that small-model PDE performance is shaped by inductive bias, not scale alone.

\section{Related Work}
\label{related_work}

\paragraph{Neural operators for PDEs.}
Neural operators approximate mappings between infinite-dimensional function spaces~\cite{kovachki2024operator}, with two main paradigms: encoder-decoder operators such as DeepONet~\cite{lu2021learning} and NoMaD~\cite{seidman2022nomad}, which map functions to latent spaces and back; and integral kernel operators~\cite{kovachki2023neural} parameterized via graph message passing~\cite{li2020neural} or Fourier transforms~\cite{li2020fourier}. These methods established that operator-valued targets can be learned end-to-end, but typically at single resolution and without explicit multi-scale structure --- the property our design is built to exploit.

\paragraph{Vision transformers for operator learning.}
ViTs~\cite{dosovitskiy2020image} have been adapted to operator learning, most prominently by CViT~\cite{wang2024cvit}, which combines ViT encoders with coordinate embeddings and cross-attention. These approaches inherit two defaults from vision: quadratic-cost softmax attention and coarse patch tokenization. The two interact: quadratic attention forces large patches to keep sequence lengths tractable, and large patches discard the fine-scale physics that PDE solutions actually depend on. WaveLiT removes both defaults simultaneously --- linear attention enables fine-grained tokens, and a wavelet transform supplies them losslessly.

\paragraph{Efficient sequence modeling.}
Linear attention~\cite{katharopoulos2020transformers} reduces the $\mathcal{O}(N^2)$ cost of softmax attention to $\mathcal{O}(N)$. Recent work has reframed such mechanisms through a test-time regression lens: DeltaNet~\cite{yang2024gated} frames the recurrent state update as gradient descent on a per-token MSE loss, and the test-time-regression perspective~\cite{wang2025test, behrouz2025s} shows that many modern sequence models can be derived from specific regression objectives. We adopt the Mamba-Inspired Linear Attention block~\cite{han2024demystify} as a starting point and develop a ridge-corrected variant motivated by this perspective (\S\ref{background}), specifically adapted to the bidirectional, spatially-tokenized setting of grid-structured PDE data -- a regime that prior linear-attention work has largely targeted only in causal sequence settings.

\paragraph{Wavelets in deep learning.}
Wavelet transforms decompose signals into components localized jointly in space and frequency, providing a principled framework for multi-resolution analysis~\cite{mallat1989theory, hernandez1996first}. In deep learning, wavelets have been used for image compression~\cite{acharya2004jpeg2000} and as a tokenization scheme for vision transformers~\cite{zhu2024wavelet}. Within operator learning, Wavelet Neural Operators perform kernel integration in the wavelet domain~\cite{tripura2022wavelet}, and Wavelet Diffusion Neural Operators apply diffusion-based modeling in the wavelet domain for PDE simulation and control~\cite{hu2024wavelet}. Our use of wavelets is distinct: we use the DWT not as the substrate for a specialized operator, but as a parameter-free tokenization that preserves multi-scale structure end-to-end through a transformer backbone, with the wavelet domain reused as an auxiliary loss to enforce spectral fidelity.

\paragraph{Foundation models for physics simulation.}
The field has shifted toward general-purpose ``foundation models'' that aim to solve a diverse range of PDE problems~\cite{nguyen2025physix, herde2024poseidon, hao2024dpot}, with recent entries reaching billions of parameters: Walrus at 1.2B~\cite{mccabe2025walrus}, PhysiX at 4.5B~\cite{nguyen2025physix}. These models achieve strong benchmark results but require infrastructure inaccessible to most research groups, and the implicit hypothesis -- that scale is what makes cross-family transfer possible -- has not been tested against architecturally efficient alternatives. Our foundation-model variant is designed precisely as such a test: at 10M parameters, memorization is mechanically infeasible, so any cross-family transfer that does occur must come from shared architectural priors.

\section{Architecture}
\label{background}

\begin{figure}
    \centering
    \includegraphics[width=\linewidth]{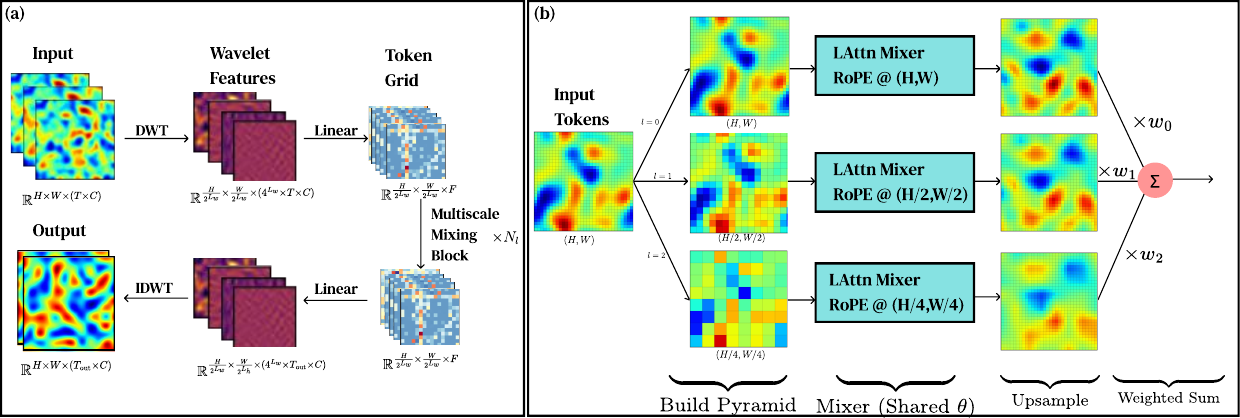}
    \caption{The \emph{WaveLiT} architecture: (a) Input fields are tokenized via a single-level 2D discrete wavelet transform (DWT), projected to embedding dimension via a linear layer, and processed by $N_L$ multiscale mixing blocks. The output is recovered by a final linear projection and inverse DWT, restoring the original spatial resolution. (b) Each multiscale mixing block applies a shared-weight linear attention mixer at $L + 1$ resolutions obtained by successive average pooling. Per-level RoPE frequencies are calibrated to each grid size; outputs are upsampled to the base resolution and combined via learned scalar weights $w_\ell$}
    \label{fig:wavelit_schematic}
    \vspace{-12pt}
\end{figure}

WaveLiT is an integrated architecture for neural PDE solvers built around four components: discrete wavelet tokenization, a gated linear attention block, a shared-weight multiscale feature pyramid, and a wavelet-domain auxiliary loss. Each component targets a specific architectural default inherited from vision transformers that we identified in Section~\ref{introduction}. We describe each in turn, then describe the foundation-model extension WaveLiT-FM. Figure~\ref{fig:wavelit_schematic} summarizes the architecture.

\subsection{Wavelet Tokenization}
We replace the standard patch embedding with a single-level 2D discrete wavelet transform (DWT). The DWT decomposes each spatial field into four half-resolution subbands (one low-frequency approximation $LL$ and three directional detail bands $LH$, $HL$, $HH$), yielding the same $4\times$ channel expansion and $2\times$ spatial reduction as a $2\times2$ patch projection but with three key advantages: the transform is (i)~\emph{lossless}, admitting exact reconstruction via the inverse DWT; (ii)~\emph{physically meaningful}, separating large-scale structure from edges and directional textures; and (iii)~\emph{parameter-free}, supplying an inductive bias with no learnable weights. For inputs of shape $B \times T \times H \times W \times C$, an $\ell$-level DWT produces tokens of shape $B \times T \times (H/2^\ell) \times (W/2^\ell) \times (C \cdot 4^\ell)$; a linear projection then maps the concatenated subbands to the model dimension. We use $\ell = 1$ throughout to preserve fine-grained features at the input layer; this choice is consistent with recent findings that smaller patch sizes outperform parameter scaling in sub-quadratic vision architectures at fixed compute~\cite{wang2025scaling}. We use the \texttt{bior2.2} wavelet throughout (ablations in Appendix~\ref{app:wavelet_selection}) and apply the inverse DWT at the output to recover the original resolution. 

In practice, wavelet tokenization yields a modest accuracy gain over a convolutional encoder-decoder (+1.4\%) while reducing training time by 11\%, consistent with the finding of \cite{hoogeboom2023simple} in a diffusion setting; the inductive bias primarily provides efficiency rather than a large accuracy jump. Full tokenizer ablations are in Appendix~\ref{app:cumulative_ablation}.

\subsection{Spatial Linear Attention Mixer}

On high-resolution grids, the $\mathcal{O}(N^2)$ cost of softmax self-attention quickly becomes prohibitive. We therefore replace quadratic attention with a kernelized linear attention operator whose cost scales linearly in the number of tokens. Let $\phi(\cdot)$ denote a positive feature map applied to queries and keys. The output at token $i$ can then be written as
\begin{align}
o_i = S \, \phi(q_i), \qquad
S = \sum_{j=1}^{N} v_j \phi(k_j)^\top,
\label{eq:linear_attn_main}
\end{align}
where the state $S \in \mathbb{R}^{d_v \times d_k}$ is accumulated once and reused across all query locations. This reordering removes the quadratic dependence on the number of tokens while preserving global token mixing. 

\paragraph{Ridge-corrected state update.}
We additionally apply a ridge-corrected variant of the state update, $S_\lambda = C(G + \lambda I)^{-1}$ with $C = \sum_j v_j \phi(k_j)^\top$ and $G = \sum_j \phi(k_j)\phi(k_j)^\top$, which can be motivated through the test-time regression view of linear attention~\cite{von2025mesanet, behrouz2025s} as incorporating the covariance structure of the kernel features. Full derivation in Appendix~\ref{app:linear_attention}; ablations (Appendix~\ref{app:ablation_mixer}) indicate this provides incremental rather than dominant gains relative to the spatial inductive biases described next.

\paragraph{Additional Spatial Inductive Biases.}
While linear attention addresses the computational bottleneck, it does not by itself provide a strong spatial inductive bias for dense grid-structured data. To remedy this, we adopt an MILA-style mixer design that augments linear attention with positional operators tailored to vision-like inputs. Concretely, we apply rotary positional embeddings \cite{su2024roformer} (RoPE) to the query/key feature maps, use conditional positional encoding (CPE) \cite{chu2021conditional} as a depthwise-convolutional positional term on the residual stream before the attention and MLP sub-blocks, and add locally enhanced positional encoding (LePE) \cite{dong2022cswin} as a depthwise-convolutional local bias on the attention output path. This combination injects both global positional structure and local spatial bias while preserving linear-time token mixing. This design is motivated by recent MILA results showing that, in vision settings, suitable positional encodings can effectively replace forget-gate-like functionality while retaining parallelizable computation \cite{han2024demystify}.

Our final mixer combines these ingredients into a single block: RoPE-equipped linear attention for efficient global token interaction, a ridge-corrected state update motivated by the regression view above, and an MILA-style block design with CPE and LePE to supply the spatial inductive bias required on high-resolution grids. In practice, we use this full configuration throughout the model. We study the effect of the individual design choices and their relative contributions in Appendix~\ref{app:ablation_mixer}.

\subsection{Multiscale Computation}
\label{sec:multiscale}

PDE solutions exhibit structure across spatial scales. Rather than only relying on network depth to capture cross-scale interactions, we process information at multiple resolutions using a feature-pyramid structure \cite{lin2017feature} with shared parameters.

Given input tokens $X \in \mathbb{R}^{B \times N \times D}$ on a grid of size $(H, W)$, we construct a resolution pyramid $\{X^{(\ell)}\}_{\ell=0}^{L}$ by successive $2\times$ average pooling. A shared-weight linear attention block $f_\theta$ is applied at each level:
\begin{equation}
Y^{(\ell)} = f_\theta(X^{(\ell)},\, g^{(\ell)}),
\label{eq:multiscale}
\end{equation}
where $g^{(\ell)}$ encodes the grid size at level $\ell$ for positional embeddings. Outputs are upsampled and aggregated via learned scalar weights:
\begin{equation}
\hat{X} = \sum_\ell w_\ell \cdot \mathrm{Upsample}(Y^{(\ell)}).
\end{equation}
Weight sharing provides multiscale sensitivity without increasing parameters; coarse levels process far fewer tokens, making additional levels computationally cheap. Concurrent work applies a similar resolution-pyramid strategy with shared attention weights for PDE simulation, achieving linear complexity via a multipole analogy~\cite{colagrande2025linear}; our design differs in using average pooling for downsampling and learned scalar weights for aggregation. A single pyramid level reduces error by 17.4\% over wavelet-only tokenization at 16\% additional training cost; additional levels yield diminishing returns. We use $L=1$ throughout; full ablations across tokenizer and FPN depth are in Appendix~\ref{app:cumulative_ablation}.

\subsection{Wavelet-Domain Auxiliary Loss}
\label{sec:wavelet_loss}

The wavelet basis used for tokenization also defines a natural multi-scale loss. In addition to a pixel-space MSE loss on predicted fields, we apply an $L_1$ loss in the wavelet domain: predictions and targets are both DWT-transformed, and the absolute error is summed across all subbands. This encourages the model to match high-frequency detail coefficients as well as low-frequency approximations, which the MSE loss alone tends to underweight. Appendix~\ref{app:loss_ablation} reports an ablation showing that the auxiliary loss reduces error power across the entire spatial-frequency spectrum.

\subsection{Foundation-Model Extension: WaveLiT-FM}
\label{sec:foundation_model}

\begin{figure}[!htpb]
\centering
\includegraphics[width=\linewidth]{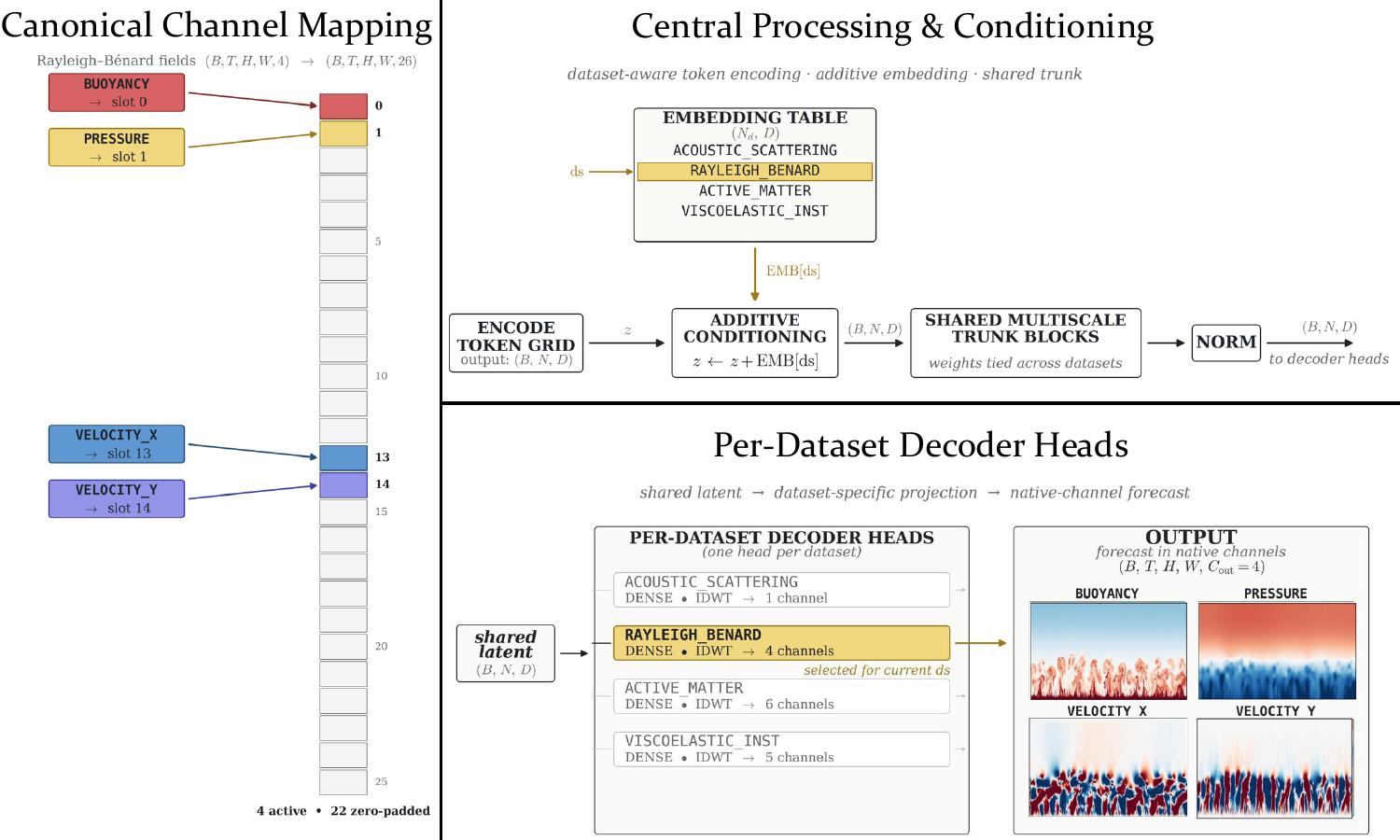}
\caption{Foundation model design: (i) input standardization via a per-channel lifting matrix without bias, so absent (zero-padded) channels contribute exactly zero to the trunk input; (ii) problem-embedding conditioning of a shared multiscale transformer trunk; and (iii) per-dataset decoder heads decoding the shared latent representation into native output channels.}
\label{fig:wavelit_fm}
\end{figure}

WaveLiT-FM extends the bespoke architecture to joint training across all eight TheWell benchmarks. The key additions over the single-task design, illustrated in Figure~\ref{fig:wavelit_fm}, are a per-channel lifting embedding that gates absent fields to zero, a per-dataset task conditioning vector, and per-dataset output heads.

\paragraph{Unified input representation.}
We define a canonical channel space of dimension $C_{\mathrm{total}}$ by taking the union of all unique physical variables (e.g., $u$-velocity, pressure, buoyancy, tracer) across TheWell. Each dataset maps its available fields to their designated canonical channels and zero-pads the rest. Channel identity is encoded through a per-channel lifting matrix $W_{\mathrm{lift}} \in \mathbb{R}^{C_{\mathrm{total}} \times T n_{\mathrm{bands}} \times D}$: after the wavelet decomposition, each channel's coefficients are projected independently and summed, $\mathbf{z} = \sum_c \mathbf{x}_c W_{\mathrm{lift}}[c]$. Because there is no bias term, absent (zero-padded) channels contribute exactly zero to the embedding --- the shared trunk receives a representation that is naturally gated by which fields are present. Channels shared across datasets (e.g., $u$-velocity in both \texttt{shear\_flow} and \texttt{rayleigh\_benard}) reuse the same rows of $W_{\mathrm{lift}}$, encouraging a common latent representation for each physical field type.

\paragraph{Task-specific conditioning.}
After the initial linear projection, we add a learnable per-dataset embedding vector to every token. This informs the shared trunk which PDE system it is solving without breaking weight sharing.

\paragraph{Shared backbone.}
The wavelet embedding, MILA stack, and reconstruction layer serve as a shared trunk trained jointly on all eight datasets; per-task specialization is confined entirely to the task-conditioning vector and output head, with no dataset-specific parameters in the core computation.

\paragraph{Specialized output heads.}
Each dataset has a separate lightweight linear head that decodes the final shared latent representation into its specific output channels. The head contributes negligible parameters relative to the trunk.

\section{Results}
\label{sec:results}

\subsection{Bespoke Models: Parameter Efficiency}
\label{sec:bespoke_results}

\paragraph{Implementation details.}
We train two model sizes (1.2M and 9.5M parameters) on each TheWell benchmark separately using AdamW~\cite{loshchilov2017decoupled} with exponential learning rate decay and gradient clipping. Both variants use a single-level DWT tokenization, the combined MSE and wavelet loss (Section~\ref{sec:wavelet_loss}), and exponential parameter averaging at evaluation. We additionally train rollout-finetuned (FT) variants by continuing from the pretrained checkpoint with an autoregressive rollout objective to improve long-horizon stability. Our rollout-finetuning protocol and the comparison that motivates it are detailed in Appendix~\ref{app:rollout_finetuning}. All training is performed in JAX~\cite{jax2018github}/Flax~\cite{flax2020github} on a single GPU; full hyperparameters are in Appendix~\ref{app:training_details}.

\begin{figure}[t]
    \centering
    \includegraphics[width=\linewidth]{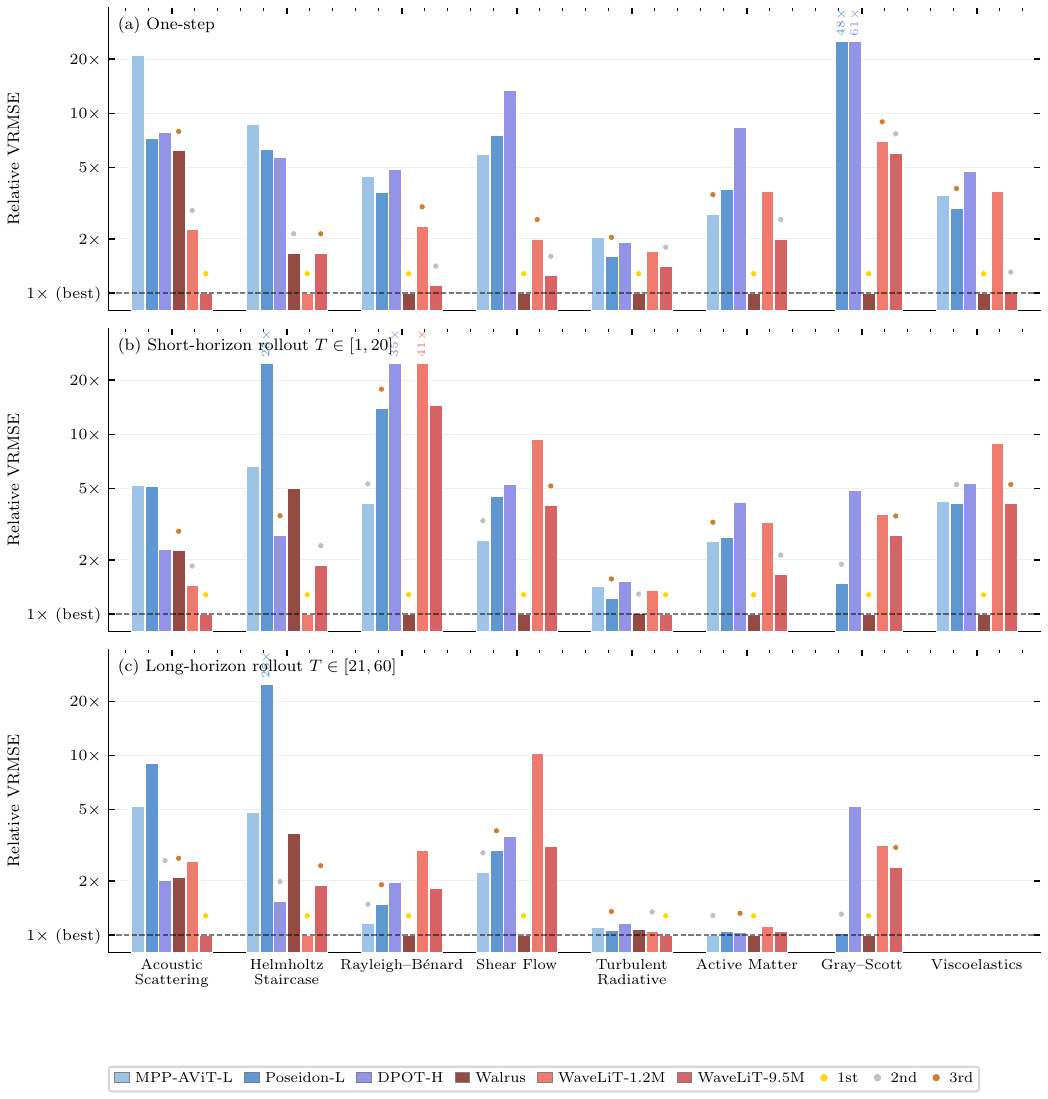}
    \caption{WaveLiT bespoke models vs.\ foundation model baselines across eight TheWell benchmarks and three evaluation windows, normalized per dataset to the best model (1$\times$ = best, log scale). Each WaveLiT bar shows the better of pretrained and rollout-finetuned variants. Gold, silver, and bronze markers denote the top three results per benchmark. Absolute VRMSE values and full PT/FT breakdown in Appendix~\ref{app:full_results}.}
    \label{fig:results_barplot}
\end{figure}

We report each model's median VRMSE normalized to the per-dataset best baseline. WaveLiT bespoke models match or exceed foundation-model baselines with 100--1000$\times$ more parameters (Figure~\ref{fig:results_barplot}; absolute VRMSE in Appendix~\ref{app:full_results}). We compare against MPP-AViT-L (409M)~\cite{mccabe2023multiple}, Poseidon-L (629M)~\cite{herde2024poseidon}, DPOT-H ($\sim$1B)~\cite{hao2024dpot}, and Walrus (1.2B)~\cite{mccabe2025walrus}.

\paragraph{One-step prediction (panel a).}
Averaged across the eight TheWell benchmarks, WaveLiT-9.5M ranks as the second-best model in the one-step regime with a mean per-dataset rank of $1.94$, trailing only Walrus ($1.44$) and outperforming all other foundation-model (FM) baselines. Remarkably, even WaveLiT-1.2M—a model $\sim$1000$\times$ smaller than Walrus—outranks the remaining three FM baselines, achieving a mean rank of $3.12$ compared to Poseidon-L ($4.12$), MPP-AViT-L ($4.86$), and DPOT-H ($5.38$) (Table~\ref{tab:avg_rank}). On the two wave- and acoustic-dominated benchmarks, ASM and HS, our architecture achieves state-of-the-art results despite its size. Specifically, WaveLiT-9.5M outperforms every baseline including Walrus on ASM ($0.0016$ vs.\ $0.0099$), and WaveLiT-1.2M achieves the same feat on HS ($0.0003$ vs.\ $0.0005$). Across the remaining six datasets, WaveLiT consistently surpasses MPP-AViT-L, Poseidon-L, and DPOT-H. The performance gap to Walrus remains narrow on RB, SF, and VI, but widens on TRL2D, AM, and GS, where Walrus's billion-parameter capacity and design likely provide a decisive advantage.

\paragraph{Longer rollouts (panels b, c).}
However, this initial one-step advantage does not carry uniformly to longer rollouts. WaveLiT-9.5M's mean rank degrades from $1.94$ at one-step to $2.38$ at $T \in [1, 20]$ and $3.00$ at $T \in [21, 60]$, with Walrus' margin widening accordingly (Table~\ref{tab:avg_rank}). When the learned map has a local Lipschitz constant $L_F > 1$, the per-step error $\epsilon$ accumulates as $\epsilon(L_F^n - 1)/(L_F - 1)$. Rollout finetuning does not reduce $\epsilon$ in isolation; rather, it trades $\epsilon$ against $L_F$. This trade-off is consistent with the worse one-step accuracy of our finetuned (FT) variants relative to their pre-trained (PT) counterparts (Appendix~\ref{app:rollout_error}). On chaotic dynamics, $L_F$ is bounded below by $e^{\lambda \Delta t}$, where $\lambda$ is the leading Lyapunov exponent of the underlying flow and $\Delta t$ is the time step; geometric error growth therefore survives finetuning. Consequently, our long-horizon wins occur on non-chaotic systems where $L_F \leq 1$, such as ASM (9.5M-FT: $0.0268$ vs.\ Walrus $0.0560$) and HS (1.2M-PT: $0.0020$ vs.\ DPOT-H $0.0031$). Conversely, evolution on RB is dominated by fine-scale structures that inherently drive $L_F > 1$, causing both our models to diverge under rollout.\footnote{The 9.5M-FT RB checkpoint uses early stopping at 40k steps; the final 50k checkpoint diverges further. The 1.2M-FT checkpoint exhibits the same instability without recovery from earlier checkpoints.}

\subsection{Foundation Model Diagnostic}
\label{sec:fm_results}

\begin{figure}[t]
    \centering
    \includegraphics[width=\linewidth]{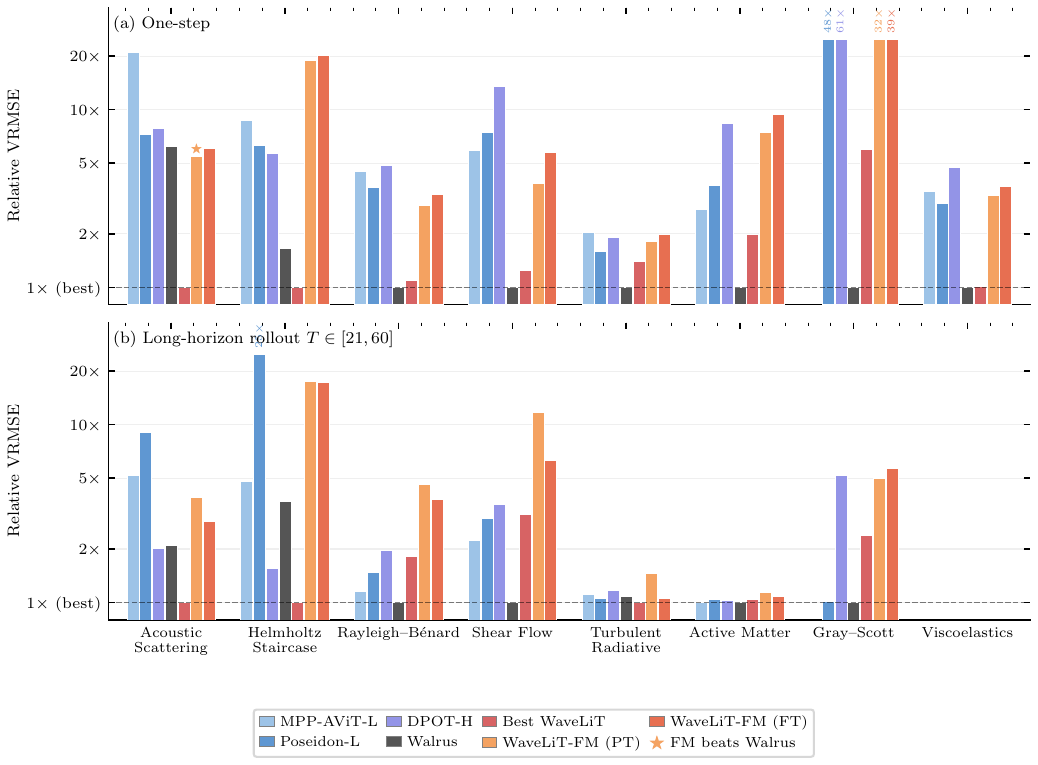}
    \caption{Foundation model diagnostic. Relative VRMSE (normalized per dataset to the best model; $1\times$ = best, log scale) for all four baselines, the best bespoke WaveLiT, and both WaveLiT-FM variants. $\bigstar$ marks the one cell where a multi-task model (Fnd-PT on ASM, one-step) outperforms Walrus.}
    \label{fig:fm_results_barplot}
\end{figure}

WaveLiT-FM is a 10M-parameter model trained jointly across all eight TheWell benchmarks on a single GPU, using the shared-backbone design of Section~\ref{sec:foundation_model} with $\sqrt{N}$-proportional dataset sampling. It is not designed to maximize benchmark performance; it is a diagnostic instrument for whether the architectural priors survive cross-family training. Figure~\ref{fig:fm_results_barplot} compares Walrus, the best bespoke WaveLiT variant, and both FM variants (pretrained and rollout-finetuned) at one-step and long-horizon rollout; full tables are in Appendix~\ref{app:full_results}.

\paragraph{Summary of Results.}
At one-step on ASM, both multi-task variants fall below Walrus ($0.0099$): Fnd-PT ($0.0087$, $\bigstar$) and Fnd-FT ($0.0097$). Together with Fnd-FT on TRL2D at long horizon ($0.8471$ vs.\ Walrus $0.8648$), these are the only three cells in this comparison where a multi-task model beats Walrus. Overall, the FM's performance gap relative to the bespoke WaveLiT ceiling is modest on most datasets (RB and SF $\sim$3$\times$, ASM and GS $\sim$5$\times$), with one pronounced exception: HS, where it widens to $19\times$.

\paragraph{Pattern of transfer.} The FM's degradation relative to bespoke is structured. HS shows by far the largest relative degradation under shared-trunk training ($19\times$ at one-step). It is the dataset where the bespoke specialist achieved its lowest absolute error ($0.0003$) by aggressively exploiting dataset-specific structure, and joint training forces a shared, generalized basis that blurs the hyper-specific filters it depends on. Conversely, datasets where bespoke and Walrus performance were already comparable (ASM, RB, TRL2D) show much smaller relative gaps. Ultimately, the FM inherits the bespoke model's strengths and weaknesses in proportion to how well the inductive bias aligns with the local dynamics. This suggests that future scaling of WaveLiT-FM should prioritize specialized per-dataset sampling for complex flows rather than merely increasing backbone capacity. Furthermore, our current joint training relies on static $\sqrt{N}$-proportional sampling, meaning the model's exposure to each dataset is fixed by volume rather than task difficulty. A promising direction for future work is the development of an adaptive sampling strategy that dynamically adjusts batch composition based on the evolution of validation loss, potentially recovering the specialization lost in standard multi-task training.

\paragraph{An observation on transfer.} A valuable outcome of running identical architectures in both regimes is the rare opportunity to directly compare their performance dynamics. Within our experimental setup, joint training across diverse PDE families did not yield positive transfer; rather, it consistently diluted specialization. Across all eight TheWell datasets, the bespoke WaveLiT specialists outperformed the multi-task FM variant, sometimes by over an order of magnitude. At this compact scale of 10M parameters, we did not anticipate strong positive transfer, as the network likely lacks the sheer capacity required to represent multiple disjoint physical manifolds simultaneously without severe interference. This raises a fascinating open question: does unlocking synergistic benefits between fundamentally different PDE families strictly require billion-parameter backbones? Currently, these mechanisms are difficult to isolate, as many pioneering PDE foundation models~\cite{mccabe2025walrus, herde2024poseidon, hao2024dpot, nguyen2025physix} are evaluated primarily as unified systems without bespoke single-task ablations of their own backbones. Our results do not settle this question, but they do raise it: the implicit assumption that multi-PDE pretraining transfers benefits to individual tasks deserves explicit experimental support.

\section{Discussion}
\label{sec:discussion}

\paragraph{Summary of findings.}
Bespoke WaveLiT models (1--10M parameters) can match or exceed foundation models up to 1000$\times$ larger, particularly on smooth problems where our inductive bias perfectly aligns with the underlying physics. On chaotic datasets, this prior improves one-step accuracy but cannot bypass the instability of teacher-forced rollouts. Crucially, our 10M-parameter WaveLiT-FM diagnostic model exhibits the exact same performance patterns as the single-task specialists. At this scale, brute-force memorization is impossible, confirming that multi-task transfer behavior is driven by the architectural prior rather than sheer scale.

\paragraph{Limitations \& Future Directions.}
Our primary limitation is autoregressive instability on chaotic dynamics, driven by fundamental mathematical bounds rather than training budgets. Under rollout, per-step error $\epsilon$ and the local Lipschitz constant $L_F$ dictate error growth: $E_n \leq \epsilon(L_F^n - 1)/(L_F - 1)$ (Appendix~\ref{app:rollout_error}). Rollout finetuning trades $\epsilon$ against $L_F$, but it cannot suppress $L_F$ below $e^{\lambda \Delta t}$, where $\lambda$ is the leading Lyapunov exponent of the system; thus, geometric growth is unavoidable. Furthermore, resolving chaotic, high-frequency subbands inherently increases $L_F$, making our fine-grained tokenization a double-edged sword. A coarser patch embedding that never represents those modes yields a smaller effective $L_F$ and slower error growth, sacrificing one-step accuracy for potentially better long-term stability. Addressing this requires migrating from static, single-step maps to architectures equipped with strong temporal mechanisms. Immediate mitigations include temporal noise injection, or ``jitter,'' as leveraged by Walrus \cite{mccabe2025walrus}. More fundamentally, incorporating temporal linear attention would allow the model to propagate a continuous hidden state, naturally regularizing the rollout trajectory. Extending this architecture to complex geometries via lifting wavelets~\cite{sweldens1998lifting} is a natural next step.

\paragraph{Conclusions.}
WaveLiT demonstrates that combining lossless multi-resolution tokenization with linear attention and shared-weight multiscale processing yields a parameter-efficient design competitive with foundation models 100--1000$\times$ larger on the PDE families where its prior matches the underlying physics. Beyond the architecture, the results provide a data point on the inductive-bias-versus-scale question: at scales where memorization is mechanically infeasible, cross-family transfer is shaped by the structure of the prior, and the pattern of failures is physically legible.

\bibliographystyle{plain}
\bibliography{refs}


\clearpage
\appendix
\section{Linear Attention and the Ridge Regression View}
\label{app:linear_attention}

\paragraph{From softmax to linear attention.}
Standard attention computes $\mathrm{Attn}(Q,K,V) = \mathrm{softmax}(QK^\top/\sqrt{d})\,V$, whose $QK^\top$ matrix is $N \times N$ and makes both memory and compute scale quadratically with sequence length. Any positive-definite kernel $\kappa(q,k)$ admits a feature-map factorization $\kappa(q,k) = \langle\phi(q),\phi(k)\rangle$; replacing the softmax with such a factorization gives
\begin{align}
o_i = \frac{\sum_j \phi(q_i)^\top\phi(k_j)\,v_j}{\sum_j \phi(q_i)^\top\phi(k_j)}.
\end{align}
Because $\phi(q_i)$ appears only as a left-multiplier, we can reorder the computation: accumulate $S = \sum_j v_j\phi(k_j)^\top \in \mathbb{R}^{d_v\times d_k}$ once, then evaluate $o_i = S\,\phi(q_i)$ for each query. This reduces complexity to $\mathcal{O}(N)$~\cite{katharopoulos2020transformers}. We follow common practice and omit the denominator, which many architectures drop without significant performance loss, giving
\begin{align}
    o_i = S\,\phi(q_i), \qquad S = \sum_{j=1}^N v_j\,\phi(k_j)^\top.
    \label{eq:linear_attn}
\end{align}

\paragraph{RNN interpretation.}
The state $S \in \mathbb{R}^{d_v \times d_k}$ accumulates key-value associations and is queried via a dot product --- it functions as an associative memory~\cite{hinton2014parallel}. Under causal (left-to-right) ordering, $S$ is updated token-by-token: $S_i = S_{i-1} + v_i\phi(k_i)^\top$, giving an exact equivalence with an RNN whose hidden state is matrix-valued. For our setting --- bidirectional spatial attention over PDE grids --- $S$ is accumulated over all tokens simultaneously, but the $\mathcal{O}(N)$ cost and fixed-size state are retained.

\subsection{The Ridge Regression View}
\label{app:ridge_view}

The accumulated state $S$ in~\eqref{eq:linear_attn} can be given an optimization interpretation. Consider the Frobenius-regularized reconstruction loss
\begin{align}
\min_S \left[\sum_{j=1}^{N} \|v_j - S\,\phi(k_j)\|^2 + \lambda \|S\|_F^2 \right].
\label{eq:ridge-obj}
\end{align}
Setting the gradient to zero yields the closed-form solution
\begin{align}
S^* = \left(\sum_j v_j\,\phi(k_j)^\top\right)\!\left(G + \lambda I\right)^{-1},
\qquad G = \sum_j \phi(k_j)\,\phi(k_j)^\top.
\end{align}
The three regularization regimes connect directly to known architectures~\cite{wang2025test, behrouz2025s, von2025mesanet}:

\begin{center}
\renewcommand{\arraystretch}{1.4}
\begin{tabular}{lll}
\toprule
\textbf{Regime} & \textbf{Solution} & \textbf{Interpretation} \\
\midrule
$\lambda \gg \|G\|$ & $\lambda^{-1}\sum_j v_j\phi(k_j)^\top$ & Vanilla linear attention~\eqref{eq:linear_attn} \\[4pt]
Finite $\lambda$     & $\left(\sum_j v_j\phi(k_j)^\top\right)(G+\lambda I)^{-1}$ & Ridge-corrected (WaveLiT) \\[4pt]
$\lambda \to 0$      & $\left(\sum_j v_j\phi(k_j)^\top\right)G^{-1}$ & Least-squares \\
\bottomrule
\end{tabular}
\end{center}

In the large-$\lambda$ regime, $(G + \lambda I)^{-1} \approx \lambda^{-1}I$ and the Gram correction vanishes --- all feature directions are treated as equally important regardless of how frequently they are queried. The scalar $\lambda^{-1}$ is absorbed into the output projection during training, leaving the functional form of vanilla linear attention~\eqref{eq:linear_attn}. Vanilla linear attention thus corresponds to \emph{ignoring the covariance structure} of the key features entirely. WaveLiT uses a finite-$\lambda$ ridge correction, incorporating the Gram matrix $G$ to account for how frequently each direction in feature space is queried. The correction adds a matrix inverse of size $d_k \times d_k$, which is cheap when $d_k \ll N$. Ablations comparing vanilla, ridge-corrected, and least-squares variants are in Appendix~\ref{app:ablation_mixer}.

\clearpage
\section{Impact of Sequence Length and Attention Mechanism on Performance and Efficiency}
\label{app:dpa_vs_mlla}

We explore the interplay between model architecture, effective sequence length (modulated by wavelet decomposition levels), and computational cost.

\paragraph{Scope of this comparison.} The study presented here isolates the choice of \emph{token-mixing primitive} (DPA versus MILA) and is intended only to motivate the move away from $\mathcal{O}(N^2)$ self-attention in long-sequence wavelet-tokenized inputs. The configurations evaluated below predate the full WaveLiT mixer and do \emph{not} include ridge regularization and gating. The accuracy numbers should therefore be read as a baseline-vs-baseline comparison of attention kernels, not as the performance of WaveLiT itself; ablations of the full mixer recipe are reported separately in Appendix~\ref{app:cumulative_ablation}. We retain this earlier comparison because the compute/accuracy trade-off it exposes between DPA and MILA is the original motivation for our linear-attention design choice and remains informative on its own.

\begin{figure}
    \centering
    \includegraphics[width=0.9\linewidth]{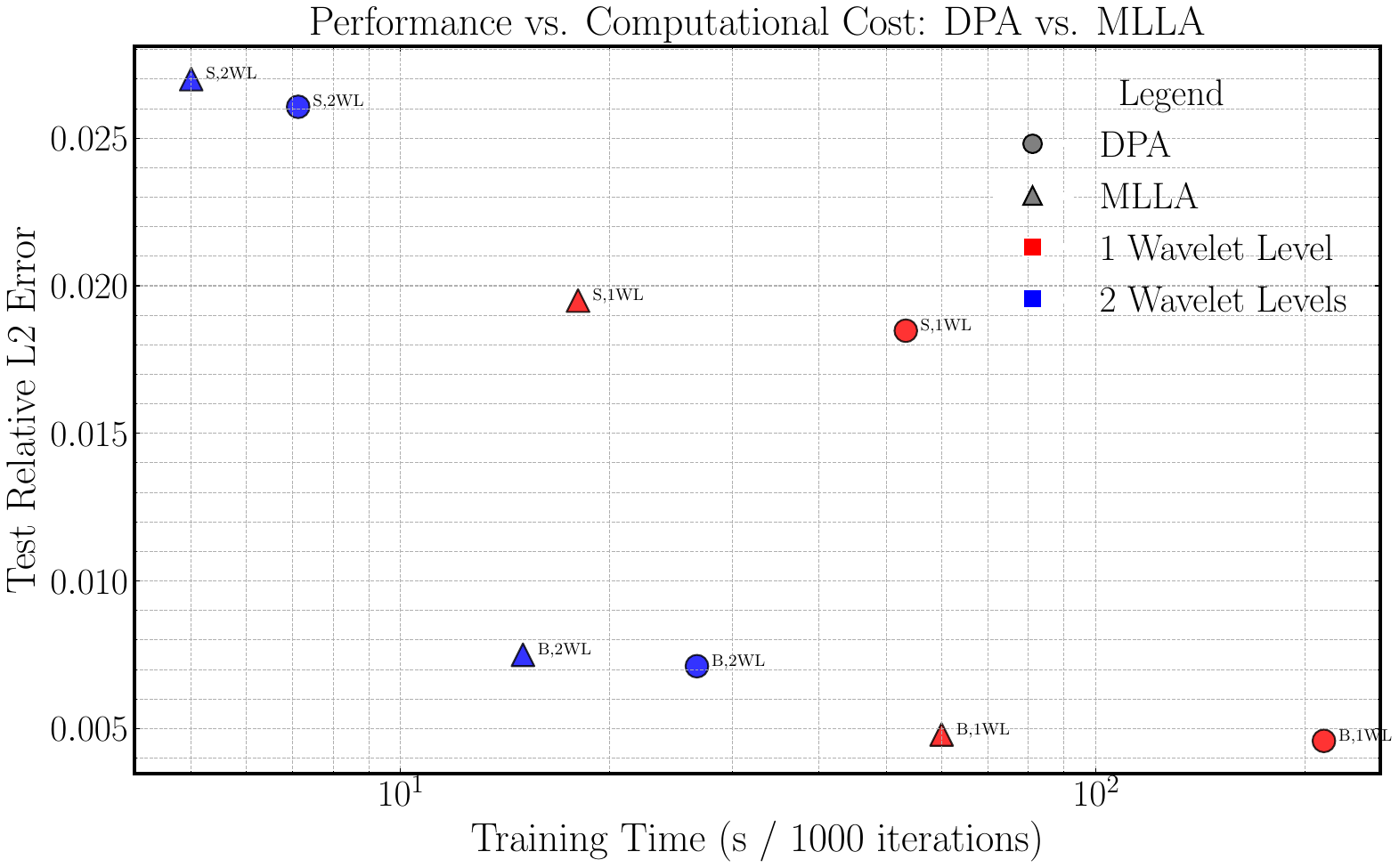}
    \caption{MILA models (triangles) demonstrate significantly lower compute costs for comparable performance levels, especially at 1 wavelet level (longer sequences), compared to DPA models (circles)}
    \label{fig:perf_vs_compute}
\end{figure}

Our primary goal is to identify model configurations that balance high accuracy with manageable computational cost. We compare standard Dot-Product Attention (DPA) with an enhanced linear attention mechanism (MILA) \cite{han2024demystify}. To understand these trade-offs, we conducted evaluations on the Navier-Stokes (NS) benchmark from PDEArena \cite{gupta2022towards}, following the experimental setup established in DPOT \cite{hao2024dpot}. We compared models using both DPA and MILA across different model sizes (``WaveLiT-1.2M'' vs.\ ``WaveLiT-9.5M'') and wavelet decomposition levels (1 vs.\ 2 levels, where fewer levels mean longer effective sequences). The results, including relative $L_2$ error and training time per 1000 iterations, are presented in Table~\ref{tab:dpa_vs_mlla_full_comparison} and visualized for key configurations in Figure~\ref{fig:perf_vs_compute}. While detailed results are shown for NS, similar trends in the DPA vs. MILA trade-off were observed across other datasets.

\begin{table}[htpb]
    \centering
    \caption{Performance and efficiency comparison of Dot-Product Attention (DPA) and Enhanced Linear Attention (MILA) across varying model sizes and wavelet levels. Training time is per 1000 iterations.}
    \label{tab:dpa_vs_mlla_full_comparison}
    \begin{tabular}{l c c c c c}
        \toprule
        Attention & Model & Wavelet Levels & Params (M) & Training Time (s) & Relative $L_2$ \\
        \midrule
        DPA       & WaveLiT-9.5M & 1              & $\approx 8.5$       & 212.80            & 0.00458     \\
        MILA      & WaveLiT-9.5M & 1              & $\approx 8.5$       & 60.00             & 0.00480     \\
        \midrule
        DPA       & WaveLiT-9.5M & 2              & $\approx 8.7$       & 26.68             & 0.00711     \\
        MILA      & WaveLiT-9.5M & 2              & $\approx 8.7$       & 15.00             & 0.00750     \\
        \midrule
        DPA       & WaveLiT-1.2M & 1              & $\approx 1.1$       & 53.30             & 0.01848     \\
        MILA      & WaveLiT-1.2M & 1              & $\approx 1.1$       & 18.00             & 0.01950     \\
        \midrule
        DPA       & WaveLiT-1.2M & 2              & $\approx 1.2$       & 7.12              & 0.02606     \\
        MILA      & WaveLiT-1.2M & 2              & $\approx 1.2$       & 5.00              & 0.02700     \\
        \bottomrule
    \end{tabular}
\end{table}

\paragraph{Discussion.}

\begin{figure}
    \centering
    \includegraphics[width=0.9\linewidth]{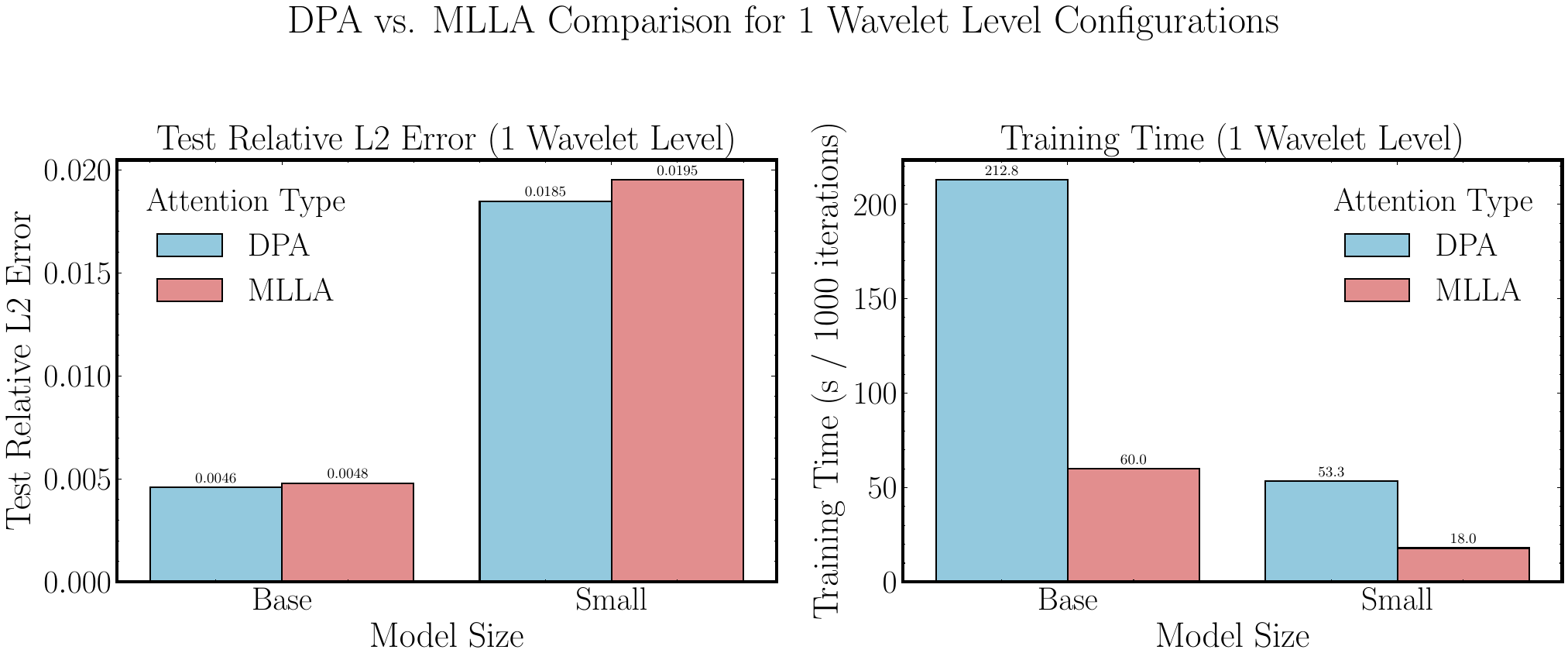}
    \caption{Comparative performance of Dot-Product Attention (DPA) and Enhanced Linear Attention (MILA) for 1-wavelet-level (long sequence) configurations. \textit{Left}: Test Relative L2 Error. \textit{Right}: Training Time (sec/1000 iterations). MILA demonstrates significant reductions in training time while maintaining competitive error rates compared to DPA across all model sizes.}
    \label{fig:bar_chart}
\end{figure}

The data in Table~\ref{tab:dpa_vs_mlla_full_comparison} (visualized in Figures~\ref{fig:perf_vs_compute} and \ref{fig:bar_chart}) reveals several key trends regarding model performance and efficiency.
Firstly, for a given model size and attention type, decreasing the number of wavelet levels (thus increasing effective sequence length) generally improves performance (lower relative $L_2$ error), albeit at a higher computational cost, especially for DPA. Secondly, increasing model parameters (e.g., from WaveLiT-1.2M to WaveLiT-9.5M) also tends to enhance performance. However, as detailed in Appendix~\ref{app:training_details}, input sequence length (determined by wavelet levels) often plays a more decisive role than raw parameter count alone, particularly for smaller models.

Most critically, these results highlight the compelling advantages of MILA. While DPA occasionally achieves marginally lower error, MILA consistently delivers comparable accuracy at a substantially reduced computational cost. For instance, with WaveLiT-9.5M and 1 wavelet level, MILA trains over 3.5 times faster than DPA for a very small trade-off in error.
Therefore, MILA emerges as a highly practical and scalable attention mechanism. It allows leveraging the performance benefits of longer effective sequences, crucial for resolving fine-scale features in PDE solutions, without incurring the prohibitive computational costs.

\clearpage
\section{Wavelet Selection Choice}
\label{app:wavelet_selection}

For the core experiments in this work, the Biorthogonal wavelet \texttt{bior2.2} was selected as the default choice for the wavelet transform layer. The \texttt{bior2.2} wavelet offers a good balance of properties, including symmetry and relatively compact support (filter length of 6), making it suitable for capturing features across various scales without excessive computational overhead. 

To validate this choice and assess the model's robustness to the specific wavelet employed, we conducted an ablation study. We evaluated the performance of the WaveLiT-1.2M model on the Navier-Stokes benchmark dataset provided by PDEArena \cite{gupta2022towards} using three distinct wavelets: \texttt{haar}, \texttt{bior2.2}, and \texttt{bior4.4}. These wavelets represent a range of characteristics: \texttt{haar} is the simplest, discontinuous orthogonal wavelet (filter length 2); \texttt{bior2.2} is our chosen symmetric biorthogonal wavelet (filter length 6); and \texttt{bior4.4} is a smoother, higher-order symmetric biorthogonal wavelet (filter length 10).  Our wavelet selection was guided by the \texttt{jax-wavelets} library \cite{Crowson_jax-wavelets}, specifically choosing from families compatible with reflect padding, which we found to yield better performance than wrap padding in our experiments.

\paragraph{Discussion.} The performance was measured using the relative-$L_2$ error for 1-step ahead predictions and 4-step rollouts. The results, presented in Table~\ref{tab:wavelet_ablation}, demonstrate  consistency across the different wavelet choices. This minimal variation in performance across wavelets with differing characteristics suggested that our architecture is reasonably robust to the specific choice of wavelet and we stick to the use of \texttt{bior2.2} as the default choice.

\begin{table}[htbp]
  \centering
  \caption{Ablation study on wavelet choice for the \texttt{WaveLiT-1.2M} model on the Navier-Stokes benchmark. Reported values are relative-$L_2$ error for 1-step ahead and 4-step rollout predictions. Lower is better.}
  \label{tab:wavelet_ablation}
  \begin{tabular}{@{}lcc@{}}
    \toprule
    Wavelet    & 1-Step Ahead & 4-Step Rollout \\
    \midrule
    \texttt{haar}   & 0.010669          & 0.023265            \\
    \texttt{bior2.2} & 0.010618          & 0.023349            \\
    \texttt{bior4.4} & 0.010896          & 0.023718            \\
    \bottomrule
  \end{tabular}
\end{table}

\clearpage
\section{Loss Term Ablations}
\label{app:loss_ablation}

Table \ref{tab:loss_term_comparison} presents the results of experiments investigating the impact of different loss term weightings on model performance, specifically comparing Mean Squared Error (MSE) loss and an $L_1$ wavelet loss ($L_1$ wavelet). The experiments were conducted for two model sizes (Small (1.2M) and Base (9.5M)) and two different numbers of wavelet decomposition levels (1 and 2).

\begin{table}[htbp]
\centering
\caption{Effect of MSE and L1 Wavelet Loss Terms on Model Performance. Performance metrics are relative $L_2$ error on the test set for single-step prediction (Test Rel. $L_2$) and 4-step rollout (Test Rel. $L_2$ Rollout). Best results for each ablation are highlighted in bold.}
\label{tab:loss_term_comparison}
\small
\begin{tabular}{@{}llcccc@{}}
\toprule
Model & Wavelet & \multicolumn{2}{c}{Loss Weights} & \multicolumn{2}{c}{Test Performance} \\
\cmidrule(lr){3-4} \cmidrule(lr){5-6}
           & Levels  & $\lambda_{MSE}$ & $\lambda_{L1}$ & Rel. $L_2$ & Rel. $L_2$ Rollout (4 steps) \\
\midrule
WaveLiT-9.5M      & 1       & 0 & 1 & \textbf{0.00416} & 0.01028 \\
WaveLiT-9.5M      & 1       & 1 & 1 & 0.00453 & \textbf{0.01009} \\
WaveLiT-9.5M      & 1       & 1 & 0 & 0.00563 & 0.01295 \\
\midrule
WaveLiT-9.5M      & 2       & 0 & 1 & \textbf{0.00662} & \textbf{0.01634} \\
WaveLiT-9.5M      & 2       & 1 & 1 & 0.00670 & 0.01636 \\
WaveLiT-9.5M      & 2       & 1 & 0 & 0.00837 & 0.01972 \\
\midrule
WaveLiT-1.2M       & 1       & 0 & 1 & 0.01177 & 0.02792 \\
WaveLiT-1.2M       & 1       & 1 & 1 & \textbf{0.01167} & \textbf{0.02702} \\
WaveLiT-1.2M       & 1       & 1 & 0 & 0.01280 & 0.02945 \\
\midrule
WaveLiT-1.2M       & 2       & 0 & 1 & 0.01407 & \textbf{0.03283} \\
WaveLiT-1.2M       & 2       & 1 & 1 & \textbf{0.01406} & 0.03295 \\
WaveLiT-1.2M       & 2       & 1 & 0 & 0.01515 & 0.03369 \\
\bottomrule
\end{tabular}
\end{table}

\paragraph{Discussion} A clear trend emerges from the data: the inclusion of the $L_1$ wavelet loss term, either exclusively ($\lambda_{L1}=1, \lambda_{MSE}=0$) or in conjunction with the MSE loss ($\lambda_{L1}=1, \lambda_{MSE}=1$), consistently yields superior performance compared to using only the MSE loss ($\lambda_{L1}=0, \lambda_{MSE}=1$).
\begin{wrapfigure}[19]{r}{0.5\columnwidth} 
    \centering
    \includegraphics[width=0.8\linewidth]{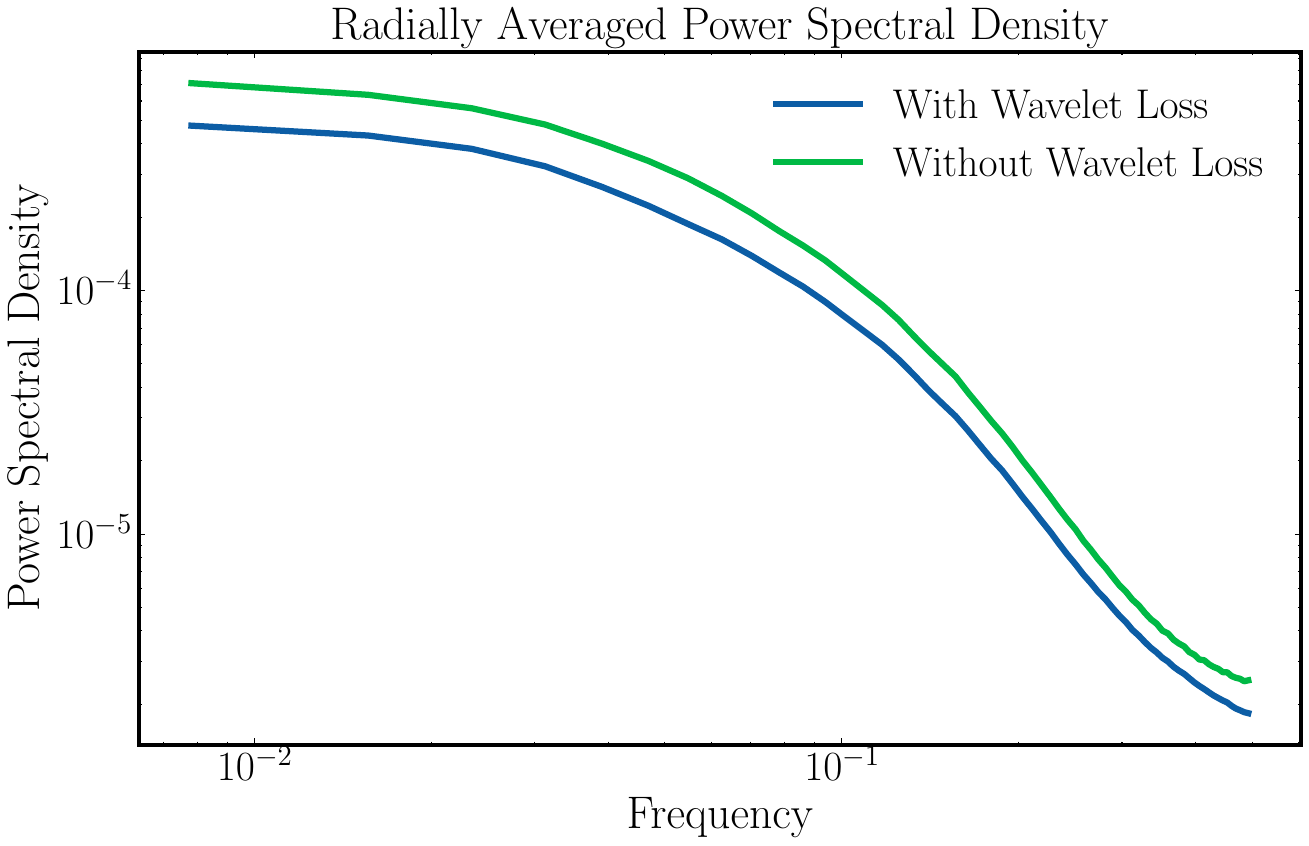}
    \caption{Radially Averaged Power Spectral Density (RAPSD) of the prediction error for the WaveLiT-9.5M model (1 wavelet level), comparing training with both MSE and $L_1$ wavelet loss terms (``With Wavelet Loss") versus training with MSE loss alone (``Without Wavelet Loss"). The inclusion of the wavelet loss demonstrably reduces error power across all frequencies.}
    \label{fig:rapsd_wavelet_loss_effect}
\end{wrapfigure}
This benefit of the wavelet loss is further demonstrated by examining the spectral characteristics of the prediction error. Figure~\ref{fig:rapsd_wavelet_loss_effect} shows the Radially Averaged Power Spectral Density (RAPSD) of the prediction error for the WaveLiT-9.5M model (1 wavelet level) when trained with the combined MSE and $L_1$ wavelet loss versus MSE loss alone. The plot indicates that the incorporation of the wavelet loss term leads to a reduction in error power across the entire frequency spectrum compared to the model trained without it. This reduction is evident at low frequencies, corresponding to large-scale structures, and extends to higher frequencies, representing finer details. Such broad spectral improvement suggests that the $L_1$ wavelet loss aids the model in more accurately resolving the multi-scale features inherent in the solution.

Given these observations, we find that setting both $\lambda_{MSE}=1$ and $\lambda_{L1}=1$ provides robust, high-quality results. Therefore, for all experiments in this paper, we adopted this balanced weighting, setting both the MSE loss weight and the $L_1$ wavelet loss weight to 1.

\clearpage
\section{Training Details}
\label{app:training_details}

All models are trained using the AdamW optimizer \citep{loshchilov2017decoupled}, using a weight decay rate of \num{1e-4}. We employ an exponential decay learning rate scheduler with a linear warm-up phase: the learning rate is linearly scaled up from \num{1e-7} to \num{1e-3} over the first 5,000 steps, after which it is exponentially decayed with a decay rate of 0.99 every 2,000 transition steps. To ensure training stability, gradients are globally clipped to a norm of 1. The batch size is set to 4 for all benchmarks.  We default to a Discrete Wavelet Transform (DWT) level of 1 for most benchmarks to maximize input sequence length and provide the model with fine-grained details.

Table~\ref{tab:training_time_bespoke} outlines the projected total training times for 500,000 steps for various WaveLiT models across different benchmarks, without intermediate evaluation. It is important to note that these times are estimates from running on a shared NVIDIA H200 server and can vary depending on factors such as data loading efficiency, software library versions, and computational facility workload; thus, they may not perfectly and consistently reflect the raw computational cost. Table ~\ref{tab:training_time_foundation} shows the timings training the foundation model backbone on an NVIDIA B200.

\begin{table}[ht]
\centering
\caption{%
  \textbf{Estimated bespoke model training time per dataset (GPU-hours, single GPU).}
  PT times estimated as $t_\text{iter} \times 500{,}000$;
  FT times as $t_\text{iter} \times 50{,}000$,
  where $t_\text{iter}$ is the mean per-iteration wall-clock time logged during training.
}
\label{tab:training_time_bespoke}
\begin{tabular}{l rr rr}
\toprule
& \multicolumn{2}{c}{\textit{Pretrain (500k iters)}}
& \multicolumn{2}{c}{\textit{Finetune (50k iters)}} \\
\cmidrule(lr){2-3} \cmidrule(lr){4-5}
\textbf{Dataset} & 1.2M & 9.5M & 1.2M & 9.5M \\
\midrule
RB  &  5.5h &  15.3h &  3.3h &  11.1h \\
ASM &  5.3h &  12.5h &  3.3h &  11.1h \\
SF  & 18.1h &  25.7h &  5.9h &  20.2h \\
AM  &  6.9h &  15.0h &  4.4h &  11.6h \\
HS  & 13.2h &  38.3h &  9.6h &  38.3h \\
GS  &  2.2h &   4.9h &  1.4h &   4.0h \\
VI  & 27.1h &  42.7h & 12.7h &  39.6h \\
TRL2D &  5.3h &   8.9h &  2.7h &   8.9h \\
\midrule
\textbf{Total} & 83.5h & 163.4h & 43.3h & 144.6h \\
\midrule
\multicolumn{5}{l}{\textit{Bespoke grand total: 434.8 GPU-hours (18.1 GPU-days)}} \\
\bottomrule
\end{tabular}
\end{table}

\begin{table}[ht]
\centering
\caption{%
  \textbf{Foundation model training time (GPU-hours, single GPU).}
  PT estimated as $t_\text{iter} \times 1{,}000{,}000$;
  FT as $t_\text{iter} \times 50{,}000$.
  Actual wall-clock time from wandb shown in parentheses for reference.
}
\label{tab:training_time_foundation}
\begin{tabular}{l rr}
\toprule
& \textit{Pretrain (1M iters)} & \textit{Finetune (50k iters)} \\
\midrule
Foundation model & 41.0h\ \ (51.1h) & 8.4h\ \ (8.5h) \\
\midrule
\textbf{Total} & \multicolumn{2}{r}{49.4 GPU-hours (2.1 GPU-days)} \\
\bottomrule
\end{tabular}
\end{table}

\begin{table}[ht]
\centering
\caption{%
  \textbf{Total compute budget summary.}
}
\label{tab:training_time_summary}
\begin{tabular}{l r r}
\toprule
\textbf{Model family} & \textbf{GPU-hours} & \textbf{GPU-days} \\
\midrule
Bespoke 1.2M (PT + FT) & 126.8h & 5.3 \\
Bespoke 9.5M (PT + FT) & 308.0h & 12.8 \\
Foundation (PT + FT)   &  49.4h &  2.1 \\
\midrule
\textbf{Grand total}   & 484.2h & 20.2 \\
\bottomrule
\end{tabular}
\end{table}

\clearpage
\section{Dataset Sampling Strategy}
\label{sec:sampling}

Training a shared-trunk foundation model across heterogeneous PDE datasets introduces
a fundamental tension: the eight datasets in our corpus differ in size by up to
\(\mathbf{66\times}\) in effective token count (Table~\ref{tab:dataset_stats}).
The trunk must accumulate useful gradient signal from every PDE family, yet
unconstrained training will concentrate update mass on whichever datasets
dominate the corpus. 

Table~\ref{tab:dataset_stats} reports, for each dataset, the number of training
trajectories, the native spatial resolution \((H \times W)\), the effective
post-wavelet token count per example (\(h/2 \times w/2\) for one decomposition
level), and the resulting total epoch token count \(N_i = n_i \cdot \tau_i\)
that serves as the basis for all sampling calculations.

\begin{table}[h]
\centering
\caption{Dataset statistics.
  $n_i$: training trajectories;
  $\tau_i$: tokens per example after one wavelet level (spatial dims halved);
  $N_i$: total epoch tokens.
  The token corpus spans nearly two orders of magnitude.}
\label{tab:dataset_stats}
\smallskip
\begin{tabular}{lR{1.0cm}R{1.4cm}R{1.7cm}R{2.2cm}}
\toprule
Dataset & $n_i$ & $H{\times}W$ & $\tau_i$ & $N_i$ (tokens) \\
\midrule
\texttt{active\_matter}                   &  14{,}000 & $256{\times}256$ &  16{,}384 & $229{,}376{,}000$ \\
\texttt{gray\_scott\_reaction\_diffusion} & 960{,}000 & $128{\times}128$ &   4{,}096 & $3{,}932{,}160{,}000$ \\
\texttt{rayleigh\_benard}                 & 278{,}600 & $512{\times}128$ &  16{,}384 & $4{,}564{,}582{,}400$ \\
\texttt{shear\_flow}                      & 178{,}304 & $256{\times}512$ &  32{,}768 & $5{,}842{,}665{,}472$ \\
\texttt{turbulent\_radiative\_layer\_2D}  &   7{,}200 & $128{\times}384$ &  12{,}288 &    $88{,}473{,}600$ \\
\texttt{viscoelastic\_instability}        &   6{,}487 & $512{\times}512$ &  65{,}536 &   $425{,}132{,}032$ \\
\texttt{acoustic\_scattering\_maze}       & 321{,}600 & $256{\times}256$ &  16{,}384 & $5{,}269{,}094{,}400$ \\
\texttt{helmholtz\_staircase}             &  20{,}384 & $1024{\times}256$ & 65{,}536 & $1{,}335{,}885{,}824$ \\
\midrule
Total & --- & --- & --- & $21{,}687{,}369{,}728$ \\
\bottomrule
\end{tabular}
\end{table}

\subsection{Three Candidate Sampling Schemes}

Let \(p_i = N_i / \sum_j N_j\) denote the proportional share of the corpus. We compare three methods for assigning the sampling probability \(w_i\) (the probability of drawing the next mini-batch from dataset~\(i\)).

\paragraph{Uniform.} $w_i = 1/K$ for all $i$, with $K=8$.

\paragraph{Proportional, temperature \(T=0.2\).}
\begin{equation}
  w_i \;\propto\; \exp\!\left(\frac{p_i}{T}\right).
  \label{eq:prop}
\end{equation}
With $T \to 0$ all weight concentrates on $\arg\max_i p_i$; with $T \to \infty$ the scheme recovers uniform. $T = 0.2$ sharpens toward the larger datasets relative to uniform. We choose this relaxation to allow sampling from sparser datasets such as \texttt{turbulent\_radiative\_layer\_2D} which would otherwise be sampled at less than 1\% under naive proportional sampling.

\paragraph{\(\sqrt{N}\)-proportional.}
\begin{equation}
  w_i \;=\; \frac{\sqrt{N_i}}{\sum_j \sqrt{N_j}}.
  \label{eq:sqrt}
\end{equation}
No temperature hyperparameter; $w_i$ lies between the pure-proportional and
uniform extremes by construction.

Table~\ref{tab:probs} lists the resulting probabilities under all three schemes alongside the raw proportional share \(p_i\) for reference. Table~\ref{tab:oversampling} provides the most informative view: the oversampling ratio \(w_i / p_i\), i.e., how many times more frequently each dataset is sampled relative to what its token share alone would dictate.

\begin{table}[h]
\centering
\caption{Sampling probabilities under each scheme.
  \emph{Prop.\ (pure)} is the raw token share $p_i$, shown for reference only.
}
\label{tab:probs}
\smallskip
\begin{tabular}{lR{1.7cm}R{1.4cm}R{1.9cm}R{1.4cm}}
\toprule
Dataset & Prop.\ (pure) & Uniform & Prop.\ ($T{=}0.2$) & $\sqrt{N}$ \\
\midrule
\texttt{active\_matter}                   & 0.0106 & 0.1250 & 0.0617 & 0.0420 \\
\texttt{gray\_scott\_rd}                  & 0.1813 & 0.1250 & 0.1448 & 0.1737 \\
\texttt{rayleigh\_benard}                 & 0.2105 & 0.1250 & 0.1676 & 0.1871 \\
\texttt{shear\_flow}                      & 0.2694 & 0.1250 & 0.2250 & 0.2117 \\
\texttt{turb.\ rad.\ layer 2D}            & 0.0041 & 0.1250 & 0.0597 & 0.0261 \\
\texttt{viscoelastic\_inst.}              & 0.0196 & 0.1250 & 0.0645 & 0.0571 \\
\texttt{acoustic\_scattering}             & 0.2430 & 0.1250 & 0.1971 & 0.2011 \\
\texttt{helmholtz\_staircase}             & 0.0616 & 0.1250 & 0.0796 & 0.1012 \\
\midrule
$D_{\mathrm{KL}}(w_i \| p_i)$ [nats]    &  0.000 &  0.766 &  0.214 & 0.099 \\
\bottomrule
\end{tabular}
\end{table}

\begin{table}[h]
\centering
\caption{Oversampling ratio $w_i / p_i$ under each scheme.
  A ratio $>1$ means the dataset is sampled more often than its raw token share
  warrants; ratio $<1$ means it is downsampled.
  Pure proportional gives ratio $1.0$ by definition (shown for reference).}
\label{tab:oversampling}
\smallskip
\begin{tabular}{lR{1.7cm}R{1.4cm}R{1.9cm}R{1.4cm}}
\toprule
Dataset & Prop.\ (pure) & Uniform & Prop.\ ($T{=}0.2$) & $\sqrt{N}$ \\
\midrule
\texttt{active\_matter}                   & $1.0\times$ & $11.8\times$ & $5.8\times$ & $4.0\times$ \\
\texttt{gray\_scott\_rd}                  & $1.0\times$ & $0.7\times$  & $0.8\times$ & $1.0\times$ \\
\texttt{rayleigh\_benard}                 & $1.0\times$ & $0.6\times$  & $0.8\times$ & $0.9\times$ \\
\texttt{shear\_flow}                      & $1.0\times$ & $0.5\times$  & $0.8\times$ & $0.8\times$ \\
\textbf{\texttt{turb.\ rad.\ layer 2D}}  & $1.0\times$ & $\mathbf{30.6\times}$ & $14.6\times$ & $\mathbf{6.4\times}$ \\
\texttt{viscoelastic\_inst.}              & $1.0\times$ & $6.4\times$  & $3.3\times$ & $2.9\times$ \\
\texttt{acoustic\_scattering}             & $1.0\times$ & $0.5\times$  & $0.8\times$ & $0.8\times$ \\
\texttt{helmholtz\_staircase}             & $1.0\times$ & $2.0\times$  & $1.3\times$ & $1.6\times$ \\
\bottomrule
\end{tabular}
\end{table}

\subsection{Analysis}
\label{sec:analysis}

We analyze the three sampling strategies based on their theoretical properties and empirical performance: Under uniform sampling, the smallest dataset (TRL2D) is drawn \(30.6\times\) more often than its token share warrants. This would cause overfitting on small datasets early in training. The temperature-scaled scheme partially corrects this, dropping the maximum oversampling ratio to \(14.6\times\). However, this requires careful tuning of the hyper-parameter $T$ and is dependent on the specific corpus distribution, requiring retuning if datasets are added or removed. We adopt \(\sqrt{N}\)-proportional sampling for the foundation model. It resolves the tensions of multi-physics training without introducing sensitive hyperparameters, offering distinct advantages:

\begin{itemize}
    \item \textbf{Minimal Distributional Distortion}: While we must smooth the sampling probabilities to prevent underfitting small datasets, we want to minimize the distortion of the natural data distribution $p_i$. Measuring the Kullback-Leibler divergence $D_{\mathrm{KL}}(w_i \| p_i)$ shows that $\sqrt{N}$ alters the true data distribution the least.
    \item \textbf{Established Precedent:} This weighting belongs to the exponential smoothing family (\(N_i^\alpha\)) established in large-scale multilingual pretraining. While models like XLM-R~\cite{conneau2020unsupervised} and mT5~\cite{xue2021mt5} tuned this exponent to \(\alpha=0.3\) to aggressively upsample the long tail, \(\sqrt{N}\) (\(\alpha=0.5\)) serves as a balanced, parameter-free middle ground.
\end{itemize}

\textbf{Empirical Validation.}
As shown in Table~\ref{tab:sampling_ablation}, we evaluated all three strategies at the 1.2M parameter scale across the full benchmark. \(\sqrt{N}\)-proportional sampling achieves the \textbf{lowest overall median VRMSE (\(2.85 \times 10^{-2}\))} and outperforms the baselines on 5 out of 8 datasets. While uniform sampling narrowly minimizes error on the smallest datasets (TRL2D and VI) due to extreme early overfitting, \(\sqrt{N}\) provides the robust, balanced signal required for generalized multi-physics training at scale.

\begin{table}[t]
\centering
\caption{Ablation of dataset sampling strategies at the 1.2M parameter
scale. VRMSE (median) on one-step prediction across eight datasets from
The Well benchmark~\cite{ohana2024well}. Best result per dataset in
\textbf{bold}.}
\label{tab:sampling_ablation}
\small
\setlength{\tabcolsep}{5pt}
\begin{tabular}{lccc}
\toprule
\textbf{Dataset}
    & \textbf{S1: Uniform}
    & \textbf{S2: Temp-Scaled ($T{=}0.2$)}
    & \textbf{S3: $\sqrt{}$-Prop (ours)} \\
\midrule
AM    & \textbf{6.07e{-}2} & 7.18e{-}2 & 7.90e{-}2 \\
GS    & 8.07e{-}3 & 6.19e{-}3 & \textbf{5.31e{-}3} \\
ASM   & 2.79e{-}2 & 1.91e{-}2 & \textbf{1.82e{-}2} \\
HS    & 8.48e{-}3 & 8.42e{-}3 & \textbf{7.23e{-}3} \\
RB    & 4.82e{-}2 & 4.15e{-}2 & \textbf{3.88e{-}2} \\
SF    & 1.27e{-}2 & 8.69e{-}3 & \textbf{8.39e{-}3} \\
TRL2D & \textbf{1.80e{-}1} & 1.96e{-}1 & 2.25e{-}1 \\
VI    & \textbf{1.79e{-}1} & 2.13e{-}1 & 2.28e{-}1 \\
\midrule
\textit{Median (overall)}
    & 3.81e{-}2 & 3.03e{-}2 & \textbf{2.85e{-}2} \\
\textit{Dataset wins}
    & 3 & 0 & \textbf{5} \\
\bottomrule
\end{tabular}
\end{table}

\clearpage
\section{Rollout Finetuning}
\label{app:rollout_finetuning}

Autoregressive models accumulate errors over long rollouts: small per-step inaccuracies compound, causing the predicted trajectory to drift increasingly far from the true solution. To address this, we explore several finetuning strategies that expose the model to its own predictions during training, thereby reducing the mismatch between training (teacher-forced) and inference regimes.

We consider four families of methods. \textbf{Scheduled Sampling} combines a 
linear teacher-forcing curriculum with $K{=}8$ BPTT unrolling: the 
teacher-forcing rate is annealed from $1.0$ to $0.1$ over the course of 
training, progressively replacing ground-truth inputs with model predictions 
while gradients are backpropagated through the rollout. \textbf{BPTT} unrolls 
the model for $K$ steps and backpropagates through the rollout without any 
teacher-forcing annealing, evaluated at $K{=}4$ and $K{=}8$. 
\textbf{CausalBPTT}~\citep{wang2022respecting} runs a fully autoregressive rollout
of $K$ steps and weights each step's loss causally:
\begin{equation}
  w_i = \exp\!\bigl(-\epsilon \textstyle\sum_{k<i} L_k\bigr),
  \label{eq:causal_weight}
\end{equation}
where $L_k$ is the loss at step $k$ (treated as a constant via
\texttt{stop\_gradient}).  At $\epsilon{=}0$ all steps are equally weighted,
recovering plain BPTT.  For $\epsilon{>}0$, step $i$ receives significant
gradient only once all earlier steps are well-predicted, implementing a
data-driven sequential curriculum without any teacher-forcing annealing. \textbf{Pushforward} unrolls $K$ steps but propagates 
gradients only through the final prediction, detaching intermediate states, 
also evaluated at $K{=}4$ and $K{=}8$.

We evaluate each method on two axes: rollout VRMSE across short ($2$:$8$), 
medium ($9$:$26$), and long ($27$:$56$) prediction horizons, and one-step 
prediction metrics to assess degradation from finetuning. The latter is 
important~---~aggressive rollout training risks degrading the model's 
single-step accuracy, which is undesirable.

Results are shown in Tables~\ref{tab:rollout} and~\ref{tab:onestep}. All 
finetuned methods improve over the baseline at longer horizons, with gains 
becoming more pronounced as rollout length increases. However, the methods 
differ substantially in how much they degrade one-step performance. Pushforward 
incurs the largest one-step penalties across both $K{=}4$ and $K{=}8$, with 
performance worsening as $K$ increases. Notably, BPTT ($K{=}8$)~---~which uses 
the same unroll length as Scheduled Sampling but without annealing~---~degrades 
one-step performance considerably, whereas Scheduled Sampling does not. This 
direct comparison isolates the effect of the teacher-forcing curriculum: gradual 
annealing allows the model to adapt to its own prediction errors without 
destabilizing single-step accuracy. Both Scheduled Sampling and CausalBPTT
emerge as strong candidates: CausalBPTT achieves competitive rollout
performance but at a measurable cost to one-step accuracy, while Scheduled
Sampling delivers comparable rollout gains with greater robustness on
single-step metrics. We therefore adopt Scheduled Sampling as our rollout
finetuning strategy for all subsequent experiments.

\begin{table}[ht]
\centering
\caption{%
  \textbf{Rollout VRMSE (median) by prediction window.}
  All results on the \textsc{TRL2D} test set. Lower is better.
  Bold indicates the best value in each column.
}
\label{tab:rollout}
\begin{tabular}{l ccc}
\toprule
\textbf{Method} & \textbf{Short [2:8]} & \textbf{Medium [9:26]} & \textbf{Long [27:56]} \\
\midrule
Baseline                              & 0.3797 & 0.7873 & 1.0239 \\
\midrule
Sched.\ Sampling ($K{=}8$)           & 0.3156 & 0.6544 & \textbf{0.8744} \\
BPTT ($K{=}4$)                        & 0.3188 & 0.7107 & 0.9310 \\
BPTT ($K{=}8$)                        & 0.3191 & 0.6457 & 0.8969 \\
CausalBPTT ($\epsilon{=}0$)           & 0.3194 & \textbf{0.6399} & 0.8774 \\
CausalBPTT ($\epsilon{=}1$)           & \textbf{0.3147} & 0.6420 & 0.8877 \\
CausalBPTT ($\epsilon{=}5$)           & 0.3169 & 0.6588 & 0.8893 \\
Pushforward ($K{=}4$)                 & 0.3508 & 0.6775 & 0.9003 \\
Pushforward ($K{=}8$)                 & 0.3713 & 0.6781 & 0.9660 \\
\bottomrule
\end{tabular}
\end{table}

\begin{table}[ht]
\centering
\caption{%
  \textbf{One-step prediction metrics.}
  All results on the \textsc{TRL2D} test set. Lower is better.
  Bold indicates the best value in each column.
}
\label{tab:onestep}
\begin{tabular}{l ccc}
\toprule
\textbf{Method} & \textbf{MSE} & \textbf{Rel.\ $\ell_2$} & \textbf{VRMSE (med.)} \\
\midrule
Baseline                              & 0.03710 & 0.1134 & 0.1466 \\
\midrule
Sched.\ Sampling ($K{=}8$)           & \textbf{0.03515} & \textbf{0.1123} & 0.1505 \\
BPTT ($K{=}4$)                        & 0.03585 & \textbf{0.1123} & 0.1487 \\
BPTT ($K{=}8$)                        & 0.03640 & 0.1162 & 0.1595 \\
CausalBPTT ($\epsilon{=}0$)           & 0.03623 & 0.1159 & 0.1585 \\
CausalBPTT ($\epsilon{=}1$)           & 0.03621 & 0.1151 & 0.1577 \\
CausalBPTT ($\epsilon{=}5$)           & 0.03560 & 0.1133 & 0.1528 \\
Pushforward ($K{=}4$)                 & 0.04257 & 0.1232 & 0.1628 \\
Pushforward ($K{=}8$)                 & 0.04900 & 0.1340 & 0.1851 \\
\bottomrule
\end{tabular}
\end{table}

\begin{figure}[ht]
\centering
\includegraphics[width=\linewidth]{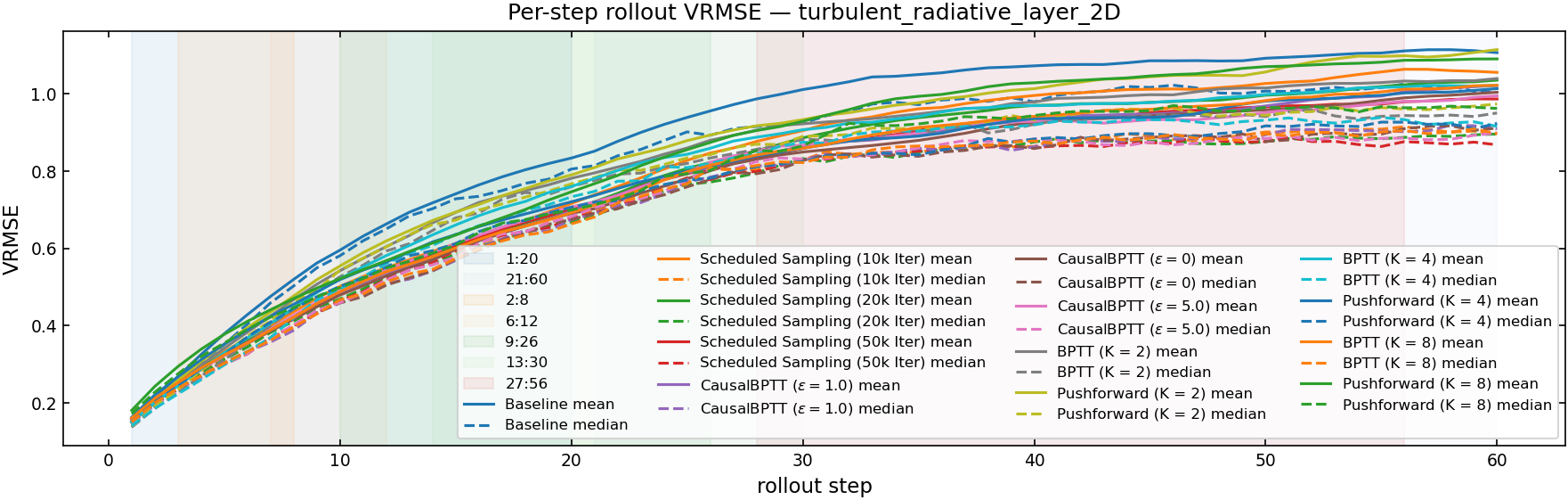}
\caption{%
  \textbf{Median rollout VRMSE as a function of prediction step (TRL2D).}
  All finetuned methods reduce error accumulation relative to the
  teacher-forced baseline, with gains widening at longer horizons.
  Pushforward degrades at both short and long steps as $K$ increases,
  while Scheduled Sampling and CausalBPTT maintain low error throughout.
  Shaded bands show the interquartile range over test trajectories.
}
\label{fig:rollout_curve}
\end{figure}
\clearpage
\section{Rollout Error Analysis}
\label{app:rollout_error}

All autoregressive surrogates trained with teacher forcing share a common failure mode under rollout. We formalize it here and connect it to WaveLiT's pattern of successes and failures across datasets.

\paragraph{Error recurrence.}
Let the ground-truth dynamics be governed by $x_{n+1} = f(x_n)$ and the learned map by $\hat{x}_{n+1} = F(\hat{x}_n)$. Starting from a perfect initial condition $x_0 = \hat{x}_0$, define $E_n := \|x_n - \hat{x}_n\|$. Suppose $F$ satisfies:
\begin{align}
\|f(x) - F(x)\| &\leq \epsilon \qquad \text{(per-step approximation error)}, \\
\|F(x) - F(y)\| &\leq L_F \|x - y\| \qquad \text{(Lipschitz stability condition)}.
\end{align}
Then:
\begin{align}
E_{n+1} &= \|f(x_n) - F(\hat{x}_n)\| \notag \\
         &\leq \|f(x_n) - F(x_n)\| + \|F(x_n) - F(\hat{x}_n)\| \notag \\
         &\leq \epsilon + L_F E_n,
\end{align}
and unrolling from $E_0 = 0$ yields
\begin{equation}
E_n \leq \epsilon \sum_{i=0}^{n-1} L_F^i =
\begin{cases}
\epsilon\,\dfrac{L_F^n - 1}{L_F - 1}, & L_F \neq 1, \\[8pt]
n\epsilon, & L_F = 1.
\end{cases}
\label{eq:rollout_bound}
\end{equation}
When $L_F > 1$, errors grow geometrically with rollout length, independent of how small $\epsilon$ is. Rollout finetuning trains on the rolled-out trajectory and so jointly optimizes $\epsilon$ and $L_F$, in practice trading them off against each other --- visible in our experiments as a degradation of one-step accuracy under FT relative to its PT counterpart, in exchange for tighter rollout error within the training horizon. The fundamental constraint is that on chaotic dynamics $L_F$ is bounded below by $e^{\lambda \Delta t}$, where $\lambda$ is the leading Lyapunov exponent of the underlying flow and $\Delta t$ is the time step: FT can move along the trade-off curve but cannot escape geometric error growth once the dynamics is genuinely chaotic. When $L_F \leq 1$ the dynamics are contractive and error grows at most linearly in $n$ --- long-horizon accuracy is then limited only by $\epsilon$, which a well-matched prior can make small.

\clearpage
\section{Architectural Ablations}
\label{app:cumulative_ablation}

This appendix collects the cumulative ablations that motivate the WaveLiT recipe. Section~\ref{app:tokenizer_fpn_ablation} isolates the contributions of the wavelet tokenizer and the multiscale feature pyramid, and Section~\ref{app:ablation_mixer} isolates the contributions of the mixer components.

\subsection{Tokenizer and Multiscale Feature Pyramid}
\label{app:tokenizer_fpn_ablation}

To isolate the contribution of each architectural choice, we perform a cumulative ablation on the Navier-Stokes benchmark from PDEArena~\cite{gupta2022towards}, training all configurations for 200k iterations under identical hyperparameters. Starting from a convolutional encoder-decoder baseline, we incrementally substitute each component and measure both accuracy and wall-clock training time. Results are summarized in Table~\ref{tab:tokenizer_fpn_ablation}.

\begin{table}[!htpb]
    \centering
    \caption{%
        Cumulative ablation on PDEArena Navier-Stokes (200k iterations).
        Each row adds one ingredient to the previous configuration.
        $\Delta$~Error denotes relative improvement over the convolutional baseline; higher is better.
        $\Delta$~Time denotes relative change in training time; negative means faster.%
    }
    \label{tab:tokenizer_fpn_ablation}
    \small
    \begin{tabular}{@{}lcccc@{}}
    \toprule
    \textbf{Configuration} & \textbf{VRMSE}~$\downarrow$ & \textbf{Time} & \textbf{$\Delta$~Error} & \textbf{$\Delta$~Time} \\
    \midrule
    Conv enc/dec (baseline)         & 0.01172 & 64.6\,min & ---      & ---    \\
    Wavelet enc/dec                 & 0.01155 & 57.4\,min & +1.4\%   & $-$11\%  \\
    \quad + Multiscale (FPN-L1)     & 0.00968 & 75.1\,min & +17.4\%  & +16\%  \\
    \quad + Multiscale (FPN-L2)     & 0.00928 & 84.1\,min & +20.8\%  & +30\%  \\
    \quad + Multiscale (FPN-L3)     & 0.00899 & 90.0\,min & +23.3\%  & +39\%  \\
    \quad + Multiscale (FPN-L4)     & 0.00897 & 97.0\,min & +23.5\%  & +50\%  \\
    \bottomrule
    \end{tabular}
    \vspace{-8pt}
\end{table}

Replacing the convolutional encoder-decoder with the wavelet tokenization scheme yields only a modest accuracy improvement (+1.4\%) while \emph{reducing} training time by 11\%. This is perhaps not surprising, since a convolutional encoder should be able to approximate wavelet-like features over the course of training; the gain we observe is consistent with the findings of \cite{hoogeboom2023simple}, who leveraged wavelet tokenizers for pixel-space diffusion models. The more substantial gain comes from introducing multiscale computation via the feature pyramid: a single pyramid level (FPN-L1) reduces error by 17.4\% relative to the baseline at a cost of 16\% additional training time, and additional levels provide diminishing returns. To keep the training-cost--accuracy frontier balanced, we adopt FPN-L1 as the default for all subsequent experiments.

\subsection{Mixer Components}
\label{app:ablation_mixer}

\begin{table}[!htbp]
\caption{
Ablation of the proposed mixer design. We keep RoPE fixed in all configurations and progressively add ridge regularization, kernel gating, local positional enhancement (LePE), conditional positional encoding (CPE), and the MILA-style block design. The largest improvements arise from LePE/CPE and the MILA-style block, while ridge regularization and kernel gating provide comparatively modest gains. Comparing B3 and C1 further shows that re-enabling the kernel gate yields only a small additional improvement over the no-gate MILA configuration.
}
\centering
\small
\setlength{\tabcolsep}{6pt}
\begin{tabular}{lccccccc}
\toprule
Model & Ridge & Kernel Gate & LePE & CPE & MILA Block & Test/Rel. $\downarrow$ & Runtime $\downarrow$ \\
\midrule
A1: RoPE-LA                    &  &  &  &  &  & 0.21906 & 36m 37s \\
A2: + Ridge                    & $\checkmark$ &  &  &  &  & 0.20256 & 39m 30s \\
A3: + Kernel Gate              & $\checkmark$ & $\checkmark$ &  &  &  & 0.19236 & 40m 38s \\
A4: + LePE                     & $\checkmark$ & $\checkmark$ & $\checkmark$ &  &  & 0.11463 & 40m 56s \\
A5: + CPE                      & $\checkmark$ & $\checkmark$ & $\checkmark$ & $\checkmark$ &  & 0.03055 & 43m 07s \\
\midrule
B1: MILA + LePE (no gate)      &  &  & $\checkmark$ &  & $\checkmark$ & 0.060258 & 40m 41s \\
B2: + Ridge (no gate)          & $\checkmark$ &  & $\checkmark$ &  & $\checkmark$ & 0.055471 & 42m 58s \\
B3: + CPE (no gate)            & $\checkmark$ &  & $\checkmark$ & $\checkmark$ & $\checkmark$ & 0.025324 & 46m 42s \\
\midrule
C1: Full model                 & $\checkmark$ & $\checkmark$ & $\checkmark$ & $\checkmark$ & $\checkmark$ & \textbf{0.024363} & 47m 25s \\
\bottomrule
\end{tabular}
\label{tab:mixer_ablation}
\end{table}

Table~\ref{tab:mixer_ablation} presents an additive ablation of our mixer design. Starting from a RoPE-equipped linear attention baseline (A1), ridge regularization and kernel gating provide only incremental improvements (A2--A3). In contrast, introducing local positional structure through LePE produces a substantial gain (A4), and adding CPE yields a further large improvement (A5), indicating that positional and locality-aware mechanisms are the dominant contributors in the standard linear-attention branch.

We next evaluate an MILA-inspired branch in which the kernel gate is disabled (B1--B3) in order to isolate the contribution of the MILA-style block design from that of the additional output modulation. This branch consistently outperforms its non-MILA counterpart at comparable levels of positional modeling, and the full no-gate MILA variant (B3) already surpasses the best non-MILA configuration (A5). These results suggest that the MILA-style block design contributes independently beyond the gains obtained from LePE and CPE alone.

Ultimately, each component contributes positively, making the full model (C1) the best overall configuration. The proposed recipe successfully combines two complementary perspectives: the ridge regression view of test-time regression, which motivates the regularization and gating choices, and standard positional and locality augmentations (LePE, CPE), which are critical for the bidirectional spatial setting. While the regression-inspired components provide consistent refinements, the dominant contributors remain the MILA block design and the positional/locality operators, consistent with prior findings in vision settings.

\clearpage
\section{Detailed Results Tables}
\label{app:full_results}

All tables below compare WaveLiT bespoke variants (1.2M and 9.5M parameters, pretrained and rollout-finetuned) against four foundation model baselines across eight TheWell benchmarks. Rows marked $\dagger$ indicate checkpoints that diverged under autoregressive rollout and are excluded from ranking. MPP-AViT-L was not evaluated on GS; VI trajectories contain only 16 rollout steps, leaving the $T\!\in\![21\!:\!60]$ window undefined for all models.

\begin{table}[ht]
\centering
\caption{%
  \textbf{One-step VRMSE} (median teacher-forcing next-step prediction error, lower is better).
  Bold = best per dataset.
  $\dagger$ = diverged checkpoint, excluded from ranking.
  ASM$^*$: one-step aggregates all sliding-window positions including trivially predictable early steps.
}
\label{tab:one_step}
\resizebox{\textwidth}{!}{%
\begin{tabular}{l cccc cccc}
\toprule
& \multicolumn{4}{c}{\textit{Baselines}} & \multicolumn{4}{c}{\textit{Ours}} \\
\cmidrule(lr){2-5} \cmidrule(lr){6-9}
\textbf{Dataset}
  & MPP-AViT-L & Poseidon-L & DPOT-H & Walrus
  & 1.2M-PT & 1.2M-FT & 9.5M-PT & 9.5M-FT \\
\midrule
ASM$^*$ & 0.0337 & 0.0116 & 0.0126 & 0.0099 & 0.0036 & 0.0037 & \textbf{0.0016} & 0.0017 \\
HS      & 0.0026 & 0.0019 & 0.0017 & 0.0005 & \textbf{0.0003} & 0.0004 & 0.0005 & 0.0005 \\
RB      & 0.0264 & 0.0215 & 0.0288 & \textbf{0.0059} & 0.0139 & 0.0159 & 0.0065 & 0.0077 \\
SF      & 0.0071 & 0.0090 & 0.0162 & \textbf{0.0012} & 0.0024 & 0.0308 & 0.0015 & 0.0023 \\
TRL2D    & 0.1707 & 0.1323 & 0.1601 & \textbf{0.0831} & 0.1421 & 0.1466 & 0.1167 & 0.1186 \\
AM      & 0.0157 & 0.0214 & 0.0476 & \textbf{0.0057} & 0.0211 & 0.0282 & 0.0114 & 0.0131 \\
GS      & ---    & 0.0048 & 0.0061 & \textbf{0.0001} & 0.0007 & 0.0008 & 0.0006 & 0.0008 \\
VI      & 0.1030 & 0.0878 & 0.1398 & \textbf{0.0295} & 0.1084 & 0.1244 & 0.0301 & 0.0336 \\
\bottomrule
\end{tabular}}
\end{table}

\begin{table}[ht]
\centering
\caption{%
  \textbf{Short-horizon rollout VRMSE, $T \in [1\!:\!20]$}
  (median over trajectories of mean autoregressive VRMSE over steps 1--20, lower is better).
  Bold = best per dataset.
  $\dagger$ = diverged checkpoint.
}
\label{tab:rollout_short}
\resizebox{\textwidth}{!}{%
\begin{tabular}{l cccc cccc}
\toprule
& \multicolumn{4}{c}{\textit{Baselines}} & \multicolumn{4}{c}{\textit{Ours}} \\
\cmidrule(lr){2-5} \cmidrule(lr){6-9}
\textbf{Dataset}
  & MPP-AViT-L & Poseidon-L & DPOT-H & Walrus
  & 1.2M-PT & 1.2M-FT & 9.5M-PT & 9.5M-FT \\
\midrule
ASM & 0.0797 & 0.0785 & 0.0350 & 0.0345 & 0.0239 & 0.0221 & 0.0158 & \textbf{0.0153} \\
HS  & 0.0053 & 0.0201 & 0.0022 & 0.0040 & \textbf{0.0008} & 0.0009 & 0.0017 & 0.0015 \\
RB  & 0.4109 & 1.3819 & 3.4868 & \textbf{0.0992} & 5.7239 & 4.0865 & 1.4417 & 1.8252 \\
SF  & 0.0377 & 0.0657 & 0.0772 & \textbf{0.0146} & 0.1375 & 0.1884 & 0.0588 & 0.0657 \\
TRL2D & 0.4796 & 0.4117 & 0.5134 & 0.3393 & 0.5684 & 0.4540 & \textbf{0.3362} & 0.3371 \\
AM  & 0.3195 & 0.3355 & 0.5272 & \textbf{0.1262} & 0.5098 & 0.4072 & 0.2096 & 0.2231 \\
GS  & ---    & 0.0411 & 0.1354 & \textbf{0.0278} & 0.1086 & 0.1002 & 0.0762 & 0.0824 \\
VI  & 0.1578 & 0.1532 & 0.1989 & \textbf{0.0373} & 0.4898 & 0.3340 & 0.1533 & 0.1651 \\
\bottomrule
\end{tabular}}
\end{table}

\begin{table}[ht]
\centering
\caption{%
  \textbf{Long-horizon rollout VRMSE, $T \in [21\!:\!60]$}
  (median over trajectories of mean autoregressive VRMSE over steps 21--60;
  steps 21--46 for HS, lower is better).
  Bold = best per dataset.
  $\dagger$ = diverged checkpoint.
  VI trajectories contain only 16 rollout steps; this window is undefined.
}
\label{tab:rollout_long}
\resizebox{\textwidth}{!}{%
\begin{tabular}{l cccc cccc}
\toprule
& \multicolumn{4}{c}{\textit{Baselines}} & \multicolumn{4}{c}{\textit{Ours}} \\
\cmidrule(lr){2-5} \cmidrule(lr){6-9}
\textbf{Dataset}
  & MPP-AViT-L & Poseidon-L & DPOT-H & Walrus
  & 1.2M-PT & 1.2M-FT & 9.5M-PT & 9.5M-FT \\
\midrule
ASM & 0.1390 & 0.2429 & 0.0543 & 0.0560 & 0.1166 & 0.0693 & 0.0406 & \textbf{0.0268} \\
HS  & 0.0096 & 0.0507 & 0.0031 & 0.0074 & \textbf{0.0020} & 0.0022 & 0.0047 & 0.0038 \\
RB  & 0.7468 & 0.9586 & 1.2664 & \textbf{0.6441} & 2.7116 & 1.9080 & 1.1690 & 1.4803 \\
SF  & 0.1814 & 0.2408 & 0.2880 & \textbf{0.0810} & 1.0732 & 0.8311 & 0.3812 & 0.2539 \\
TRL2D & 0.8879 & 0.8441 & 0.9316 & 0.8648 & 1.0853 & 0.8398 & 0.8759 & \textbf{0.7986} \\
AM  & 1.2481 & 1.3015 & 1.2867 & \textbf{1.2451} & 1.4186 & 1.3863 & 1.3016 & 1.3204 \\
GS  & ---    & 0.1295 & 0.6567 & \textbf{0.1268} & 0.4045 & 0.3996 & 0.3040 & 0.3048 \\
VI  & ---    & ---    & ---    & ---             & ---    & ---    & ---    & ---    \\
\bottomrule
\end{tabular}}
\end{table}

\begin{table}[ht]
\centering
\caption{%
  \textbf{Average per-dataset rank} across the eight TheWell benchmarks (lower is better).
  Each model's rank is computed within each dataset; ties get the mean rank; undefined cells are excluded (MPP-AViT-L on GS for all windows; VI for long-horizon).
  WaveLiT rows use the better of pretrained and rollout-finetuned variants per cell, matching Figure~\ref{fig:results_barplot}.
  WaveLiT-9.5M is the runner-up to Walrus in every window despite being $\sim$130$\times$ smaller.
}
\label{tab:avg_rank}
\small
\begin{tabular}{l r ccc}
\toprule
\textbf{Model} & \textbf{Params} & \textbf{One-step} & $\boldsymbol{T \in [1,20]}$ & $\boldsymbol{T \in [21,60]}$ \\
\midrule
Walrus            & 1.2B          & \textbf{1.44} & \textbf{1.75} & \textbf{2.14} \\
WaveLiT-9.5M      & 9.5M          & 1.94          & 2.38          & 3.00          \\
WaveLiT-1.2M      & 1.2M          & 3.12          & 4.25          & 4.14          \\
Poseidon-L        & 629M          & 4.12          & 3.62          & 3.86          \\
MPP-AViT-L        & 409M          & 4.86          & 3.86          & 3.50          \\
DPOT-H            & $\sim$1B      & 5.38          & 4.88          & 4.00          \\
\bottomrule
\end{tabular}
\end{table}


\subsection*{Foundation Model: Detailed Comparison}

Tables~\ref{tab:fnd_one_step}--\ref{tab:fnd_rollout_long} compare WaveLiT-FM against Walrus and the best bespoke WaveLiT variant per cell. The \emph{Best WaveLiT} column reports the minimum across all four bespoke variants (1.2M-PT/FT, 9.5M-PT/FT). $\ddagger$ marks cells where a foundation model variant beats Walrus. $\dagger$ marks rollout divergence.

\begin{table}[ht]
\centering
\caption{%
  \textbf{One-step VRMSE} — foundation model results in context.
  Bold = best per dataset across all four columns.
  $\ddagger$ = foundation model beats Walrus.
}
\label{tab:fnd_one_step}
\resizebox{0.72\textwidth}{!}{%
\begin{tabular}{l cccc}
\toprule
\textbf{Dataset}
  & Walrus & Best WaveLiT & Fnd-PT & Fnd-FT \\
  & (external) & (bespoke) & (multi-task) & (multi-task + FT) \\
\midrule
ASM & 0.0099          & \textbf{0.0016} & $0.0087^\ddagger$ & $0.0097^\ddagger$ \\
HS  & 0.0005          & \textbf{0.0003} & 0.0057            & 0.0061 \\
RB  & 0.0059          & \textbf{0.0065} & 0.0171            & 0.0198 \\
SF  & \textbf{0.0012} & 0.0015          & 0.0046            & 0.0069 \\
TRL2D & \textbf{0.0831} & 0.1167          & 0.1508            & 0.1655 \\
AM  & \textbf{0.0057} & 0.0114          & 0.0423            & 0.0540 \\
GS  & \textbf{0.0001} & 0.0006          & 0.0032            & 0.0039 \\
VI  & \textbf{0.0295} & 0.0301          & 0.0970            & 0.1100 \\
\bottomrule
\end{tabular}}
\end{table}

\begin{table}[ht]
\centering
\caption{%
  \textbf{Short-horizon rollout VRMSE, $T \in [1\!:\!20]$}.
  $\dagger$ = Fnd-PT diverges on RB under autoregressive rollout (value shown for completeness, excluded from ranking).
}
\label{tab:fnd_rollout_short}
\resizebox{0.72\textwidth}{!}{%
\begin{tabular}{l cccc}
\toprule
\textbf{Dataset}
  & Walrus & Best WaveLiT & Fnd-PT & Fnd-FT \\
  & (external) & (bespoke) & (multi-task) & (multi-task + FT) \\
\midrule
ASM & 0.0345          & \textbf{0.0153} & 0.0429  & 0.0494  \\
HS  & 0.0040          & \textbf{0.0008} & 0.0179  & 0.0183  \\
RB  & \textbf{0.0992} & 1.4417          & $5.7242^\dagger$ & $4.6113^\dagger$ \\
SF  & \textbf{0.0146} & 0.0588          & 0.1376  & 0.1091  \\
TRL2D & 0.3393          & \textbf{0.3362} & 0.6729  & 0.5181  \\
AM  & \textbf{0.1262} & 0.2096          & 0.7456  & 0.7195  \\
GS  & \textbf{0.0278} & 0.0762          & 0.1964  & 0.2222  \\
VI  & \textbf{0.0373} & 0.1533          & 0.3497  & 0.2852  \\
\bottomrule
\end{tabular}}
\end{table}

\begin{table}[ht]
\centering
\caption{%
  \textbf{Long-horizon rollout VRMSE, $T \in [21\!:\!60]$}.
  VI trajectories contain only 16 rollout steps; this window is undefined ($---$).
  $\ddagger$ = foundation model beats Walrus.
}
\label{tab:fnd_rollout_long}
\resizebox{0.72\textwidth}{!}{%
\begin{tabular}{l cccc}
\toprule
\textbf{Dataset}
  & Walrus & Best WaveLiT & Fnd-PT & Fnd-FT \\
  & (external) & (bespoke) & (multi-task) & (multi-task + FT) \\
\midrule
ASM & 0.0560          & \textbf{0.0268} & 0.1050              & 0.0765 \\
HS  & 0.0074          & \textbf{0.0020} & 0.0349              & 0.0346            \\
RB  & \textbf{0.6441} & 1.1690          & 2.9770              & 2.4407            \\
SF  & \textbf{0.0810} & 0.2539          & 0.9487              & 0.5143            \\
TRL2D & 0.8648          & \textbf{0.7986} & 1.1706              & $0.8471^\ddagger$ \\
AM  & \textbf{1.2451} & 1.3016          & 1.4208              & 1.3440            \\
GS  & \textbf{0.1268} & 0.3040          & 0.6371              & 0.7173            \\
VI  & ---             & ---             & ---                 & ---               \\
\bottomrule
\end{tabular}}
\end{table}

\begin{table}[h]
\centering
\caption{Rank-1 finishes across 20 dataset--window pairs
  (among \{Walrus, MPP-AViT-L, Poseidon-L, DPOT-H, Fnd-PT, Fnd-FT\};
  GS excluded entirely since MPP-AViT-L is not evaluated on it, and VI excluded at $T\!\in\![21\!:\!60]$).}
\label{tab:fnd_wins}
\smallskip
\begin{tabular}{lc}
\toprule
Model & Rank-1 count (of 20) \\
\midrule
Walrus  (1.2B, bespoke)          & 15 \\
Fnd-FT  (10M, multi-task + FT)   &  0 \\
Fnd-PT  (10M, multi-task)        &  1 \\
DPOT-H                           &  3 \\
Poseidon-L                       &  1 \\
MPP-AViT-L                       &  0 \\
\bottomrule
\end{tabular}
\end{table}

\clearpage
\section{Comparison with Wavelet Neural Operator}
\label{app:wno_comparison}

Wavelet-based methods for neural PDE solvers span three distinct families, and it is worth clarifying how WaveLiT relates to each before presenting comparisons:

\begin{itemize}
    \item \textbf{WNO}~\cite{tripura2022wavelet} is an integral kernel method that performs kernel integration directly in the wavelet domain --- the wavelet transform defines the space in which convolution is computed, analogous to how FNO operates in the Fourier domain.
    \item \textbf{WDNO}~\cite{hu2024wavelet} is a diffusion-based generative model that learns to denoise PDE trajectories in the wavelet domain.
    \item \textbf{WaveLiT} uses the DWT as a parameter-free tokenization scheme that feeds a transformer backbone, and reuses the wavelet domain as an auxiliary loss to enforce spectral fidelity. The wavelet transform is not the computational substrate of the operator; it is the input representation.
\end{itemize}

A direct comparison is therefore not perfectly fair, but it is nonetheless informative. We run WNO using the official repository\footnote{\url{https://github.com/TapasTripura/WNO/tree/main/Version\%202.0.0}} with default hyperparameters, adapting only the output dimensionality and spatial resolution to match each benchmark. This yields a WNO model of approximately 9M parameters, directly comparable in size to WaveLiT-9.5M.

\begin{table}[h]
\centering
\caption{One-step VRMSE comparison between WNO ($\sim$9M parameters) and WaveLiT-9.5M on two TheWell benchmarks. Lower is better. WNO uses default repository settings; WaveLiT-9.5M uses the pretrained checkpoint.}
\label{tab:wno_comparison}
\begin{tabular}{lcc}
\toprule
\textbf{Benchmark} & \textbf{WNO ($\sim$9M)} & \textbf{WaveLiT-9.5M} \\
\midrule
Active Matter           & 0.5736 & \textbf{0.0114} \\
Turbulent Radiative Layer & 0.3939 & \textbf{0.1167} \\
\bottomrule
\end{tabular}
\end{table}

WaveLiT-9.5M outperforms WNO by a large margin on both benchmarks despite identical parameter counts.


\end{document}